\documentclass[letterpaper, 10 pt, conference]{ieeeconf}  % Comment this line out if you need a4paper

\IEEEoverridecommandlockouts                              % This command is only needed if 
\usepackage{fdsymbol}
\usepackage[percent]{overpic}
\usepackage{physics}
\usepackage{array}
\usepackage{xcolor}
\usepackage{subcaption}

\newcolumntype{P}[1]{>{\centering\arraybackslash}p{#1}}
\newcommand\tstrut{\rule{0pt}{2.4ex}}

\title{\LARGE \bf
Autoencoder Based Inter-Vehicle Generalization \\ for In-Cabin Occupant Classification
}

\author{Steve Dias Da Cruz\,$^{1,2,3}$, Bertram Taetz\,$^{3}$, Oliver Wasenm\"uller\,$^{4}$, Thomas Stifter\,$^{1}$, Didier Stricker\,$^{2,3}$% <-this % stops a space
\thanks{$^{1}$IEE S.A., Luxembourg, {\{steve.dias-da-cruz, thomas.stifter\}@iee.lu}}%
\thanks{$^{2}$University of Kaiserslautern, Germany}%
\thanks{$^{3}$DFKI - German Research Center for Artificial Intelligence, Augmented Vision research department, {\{bertram.taetz, didier.stricker\}@dfki.de}}%
\thanks{$^{4}$Mannheim University of Applied Sciences, Germany, 
{o.wasenmueller@hs-mannheim.de}}%
}

\begin{document}

\maketitle
\thispagestyle{empty}
\pagestyle{empty}

%%%%%%%%%%%%%%%%%%%%%%%%%%%%%%%%%%%%%%%%%%%%%%%%%%%%%%%%%%%%%%%%%%%%%%%%%%%%%%%%
\begin{abstract}

Common domain shift problem formulations consider the integration of multiple source domains, or the target domain during training. Regarding the generalization of machine learning models between different car interiors, we formulate the criterion of training in a single vehicle: without access to the target distribution of the vehicle the model would be deployed to, neither with access to multiple vehicles during training. We performed an investigation on the SVIRO dataset for occupant classification on the rear bench and propose an autoencoder based approach to improve the transferability. The autoencoder is on par with commonly used classification models when trained from scratch and sometimes out-performs models pre-trained on a large amount of data. Moreover, the autoencoder can transform images from unknown vehicles into the vehicle it was trained on. These results are corroborated by an evaluation on real infrared images from two vehicle interiors.

\end{abstract}

%%%%%%%%%%%%%%%%%%%%%%%%%%%%%%%%%%%%%%%%%%%%%%%%%%%%%%%%%%%%%%%%%%%%%%%%%%%%%%%%
\section{INTRODUCTION}
The deployment of deep learning based approaches in the automotive industry needs to be corroborated by robustness and generalization guarantees, especially when the models' predictions would be used for safety critical applications, e.g. the adjustment of the strength of the air-bag deployment in case of an accident \cite{airbag,perrett2016cost}. In this work, we focus on camera-based occupant classification on the rear bench. We will highlight some of the unique challenges for the vehicle interior regarding the robustness and generalization of machine learning models. Machine learning models trained in a single vehicle interior take non-relevant characteristics of the background into account for their decision taking \cite{tian2018eliminating}, because the training data contains similar backgrounds for all images. Consequently, the performance drops drastically if models trained in a single car interior are used in a different vehicle. Repeating the data recording and annotation generation process for each new car model and automotive manufacturer implies a time-consuming and costly development pipeline. Alternative sensor data (e.g. depth maps computed by time-of-flight sensors, RADAR \cite{da2019theoretical}) and the inclusion of data from different vehicles (e.g. domain generalization) would improve the transferability. However, improving the theoretical foundations of deep learning models is paramount for safety critical applications. As illustrated in Fig. \ref{fig:teaser}, we formulate the challenge of training in a single vehicle interior and generalizing to unseen cars without using images from the target distribution or from multiple vehicles. Even more difficult, the models' generalization to new objects in new environments should be examined as well. This can be considered as an extreme form of domain adaptation \cite{pan2009survey} and is related to (generalized) zero-shot learning \cite{xian2017zero, chao2016empirical}. We will adopt the SVIRO dataset \cite{DiasDaCruz2020SVIRO} to investigate the generalization between different vehicle interiors. Our key contributions can be summarized as follows: 
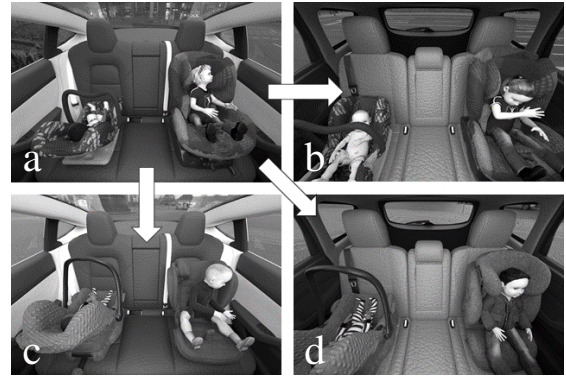
\begin{figure}
	\centering
	\begin{overpic}[width=0.85\linewidth]{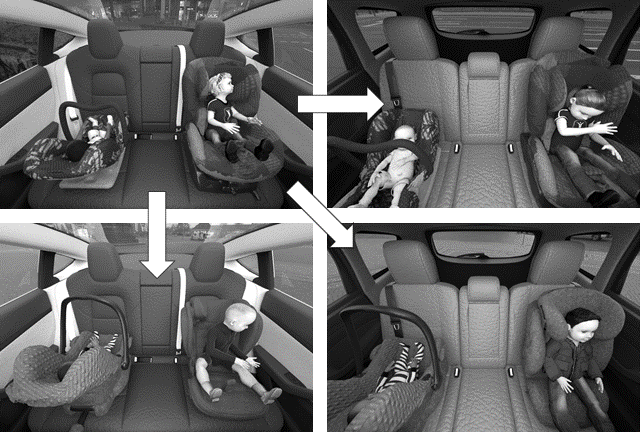}
		\put(2,37){\LARGE\textcolor{white}{a}}
		\put(53,37){\LARGE\textcolor{white}{b}}
		\put(2,2){\LARGE\textcolor{white}{c}}
		\put(53,2){\LARGE\textcolor{white}{d}}
	\end{overpic}
	\caption{Synthetic images from SVIRO \cite{DiasDaCruz2020SVIRO}. Left: infant seat with a baby, Right: child seat with a child. Training on images from a \textbf{\textit{single}} vehicle interior (a): How can we improve robustness on same class instances, but to a new vehicle (b)? Can the model generalize to new class instances in the vehicle it was trained on (c)? How can we improve generalization to new class instances in a new vehicle (d)? }
	\label{fig:teaser}
\end{figure}
\begin{itemize}
	\itemsep0.0em
	\item We demonstrate that transfer learning is not sufficient to ensure consistency between different vehicles for the aforementioned challenging training conditions, 
	\item We show that autoencoders with a classifier in the latent space achieve accuracies on par with classification models. Moreover, the autoencoders can transform images from unknown vehicles to the one they were trained on, 
	\item We compare different reconstruction cost functions in autoencoders which lead to different behaviours for classification and reconstruction transferability,
	\item  We corroborate the inter-vehicle transformation by showing that it works on real infrared images as well, even when trained on synthetic images. 
\end{itemize}

\noindent The resulting advantages are two-fold: we need less data to achieve similar performances, which is important when pre-trained models cannot be used due to licensing constraints. Future work can exploit progress in disentanglement and the inclusion of prior knowledge in the latent space of autoencoders to further improve the robustness and transferability.

%%%%%%%%%%%%%%%%%%%%%%%%%%%%%%%%%%%%%%%%%%%%%%%%%%%%%%%%%%%%%%%%%%%%%%%%%%%%%%%%

\section{RELATED WORK}
\label{section:related}
\noindent\textbf{Datasets}: Publicly available realistic datasets for the vehicle interior are scarce. Some exceptions are the recently released AutoPOSE \cite{autopose} and DriveAHead \cite{schwarz2017driveahead} datasets for driver head orientation and position, Drive\&Act \cite{drive_and_act_2019_iccv} a multi modal benchmark for action recognition in automated vehicles and TICaM \cite{katrolia2021ticam} for activity recognition and person detection. However, these datasets all have in common that they provide images for a single vehicle interior such that the transferability between vehicles cannot be tested reliably. Although SVIRO \cite{DiasDaCruz2020SVIRO} is a synthetic dataset, it was designed specifically to test the transferability between different vehicles across multiple tasks. The applicability to real infrared  \cite{DiasDaCruz2020SVIRO} and depth images \cite{pub10767} was shown. In Section \ref{section:real}, we will present that insights on SVIRO are transferable to real infrared images. Training on SVIRO and applying the resulting model to real images is possible. Moreover, recent studies have shown the importance of synthetic data for the automotive industry \cite{nowruzi2019much,tremblay2018training,chen2019learning}. 
\\

\noindent\textbf{Domain shift}: Methods from domain adaptation \cite{pan2009survey, reiss2020deep, visapp21} are commonly used to reduce the gap between the target domain (the vehicle in which we want to use our model) and source domain (the vehicle we trained on). However, these methods usually require (often even labelled) images from the target distribution to work well. Zero-shot learning (ZSL) \cite{xian2017zero, norouzi2013zero}, and particularly generalized zero-shot learning (GZSL) \cite{chao2016empirical, xian2017zero, frome2013devise}, are the most extreme cases of domain adaptation as they do not require labels for new test objects. Both setups consider the generalization to new classes, but require some additional type of information, e.g. word embeddings \cite{norouzi2013zero} or semantic descriptions \cite{frome2013devise}. However, we focus on the evaluation of \textit{seen} class \textit{instances} in \textit{unknown} environments and \textit{unknown} class \textit{instances} in \textit{known} and \textit{unknown} environments. Nevertheless, our problem setup is closest to (G)ZSL and they share some characteristics such that advances for the latter might be useful as well. The main difference stems from the following constraint: the adaptation of trained models to new class instances and environments should be avoided. For example, models should be robust against new child seats appearing on the market after the model was deployed, and models should not need to be adapted for each vehicle interior variation.

Alternatively, it would be possible to transform images from an unknown vehicle back to the known vehicle, e.g. by aligning both domains \cite{ liu2017unsupervised} or by using style transfer techniques \cite{karras2019style, tsai2018learning}. However, those techniques need images from the target distribution as well. Similar to eyeglass removal achieved by generative adversarial networks (GAN) \cite{rangesh2020driver}, we could use a GAN to change the vehicle background, but this would need images from the target domain or image-pairs of what we would expect to encounter. Domain generalization considers methods to generalize to new domains without accessing images from the unknown domain during training \cite{li2017deeper, zhou2020learning}. Nevertheless, these techniques use images from several domains during training to learn generalizing well to unknown domains and the aforementioned methods often combine several datasets. To the best of our knowledge, SVIRO is the first dataset which allows to investigate the generalization on the same tasks to a new, but similar, vehicle interior. Hence, common domain shift problem formulations and proposed solutions could not consider the challenge of generalizing to an unknown domain when learning from a single domain for solving the same task. 
\\

\noindent\textbf{Representation learning}: Recent advances have shown that disentanglement in the latent space of autoencoders can lead to improved performance on visual downstream tasks \cite{VanSteenkiste2019}. It is believed that meaningful scene decomposition \cite{burgess2019monet,engelcke2019genesis} will improve the transferability for many tasks, however, these methods still do not work well for scenes of higher visual complexity and are currently limited to toy datasets like CLEVR \cite{johnson2017clevr}, Objects Room \cite{burgess2019monet} or MPI3D \cite{Gondal2019}. We provide a baseline for the transferability between vehicle interiors with our proposed autoencoder approach. Hence, future work can analyse the effect of disentanglement and scene decomposition with respect to the transferability on a task with higher visual complexity than commonly used datasets. These advances can be compared against self-supervised learning methods \cite{chen2020simple, noroozi2016unsupervised}, for example, with respect to resulting differences in the latent space.

%%%%%%%%%%%%%%%%%%%%%%%%%%%%%%%%%%%%%%%%%%%%%%%%%%%%%%%%%%%%%%%%%%%%%%%%%%%%%%%%

\section{METHOD}
\label{section:method}
\begin{figure}
	\centering
	\begin{overpic}[width=\linewidth]{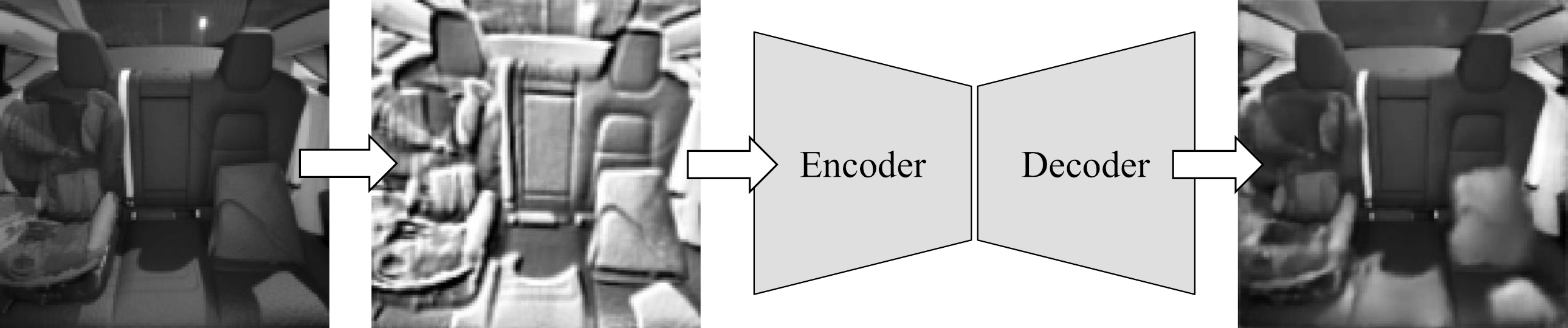}
		\put(1.5,2){\LARGE\textcolor{white}{a}}
		\put(25.5,2){\LARGE\textcolor{white}{b}}
		\put(80.5,2){\LARGE\textcolor{white}{c}}
	\end{overpic}
	\caption{Image (a) from the SVIRO dataset is augmented using different random transformations to form image (b). The latter is used as input to the autoencoder, which should learn to output image (c), i.e. transform (b) back to (a).}
	\label{fig:augmentation}
\end{figure}

Our method is based on exploiting the disadvantage of our problem formulation to our advantage. As all the images from a single vehicle interior contain similar backgrounds, an autoencoder should be able to learn to reconstruct the vehicle interior robustly. If we augment the training images, then the autoencoder should easily learn to clean the augmented images, because the clean backgrounds do not undergo a high variability. Example images are shown in Fig. \ref{fig:augmentation}: we augment the training images (a) to obtain a modified version (b) by changing colours and the perspective and adding random noise. The autoencoder then reconstructs (c), which should be the initial clean version (a). Similar approaches have been applied to de-noising and in-painting \cite{xie2012image,vincent2010stacked}. The autoencoder needs to map different backgrounds and perspectives to similar latent space representations to reconstruct the background correctly. Hence, a classifier using the latent space vector as input should learn to neglect background information more easily than using the input image. 

\subsection{Architecture Details}
\begin{figure}
	\centering
	\includegraphics[width=0.80\linewidth]{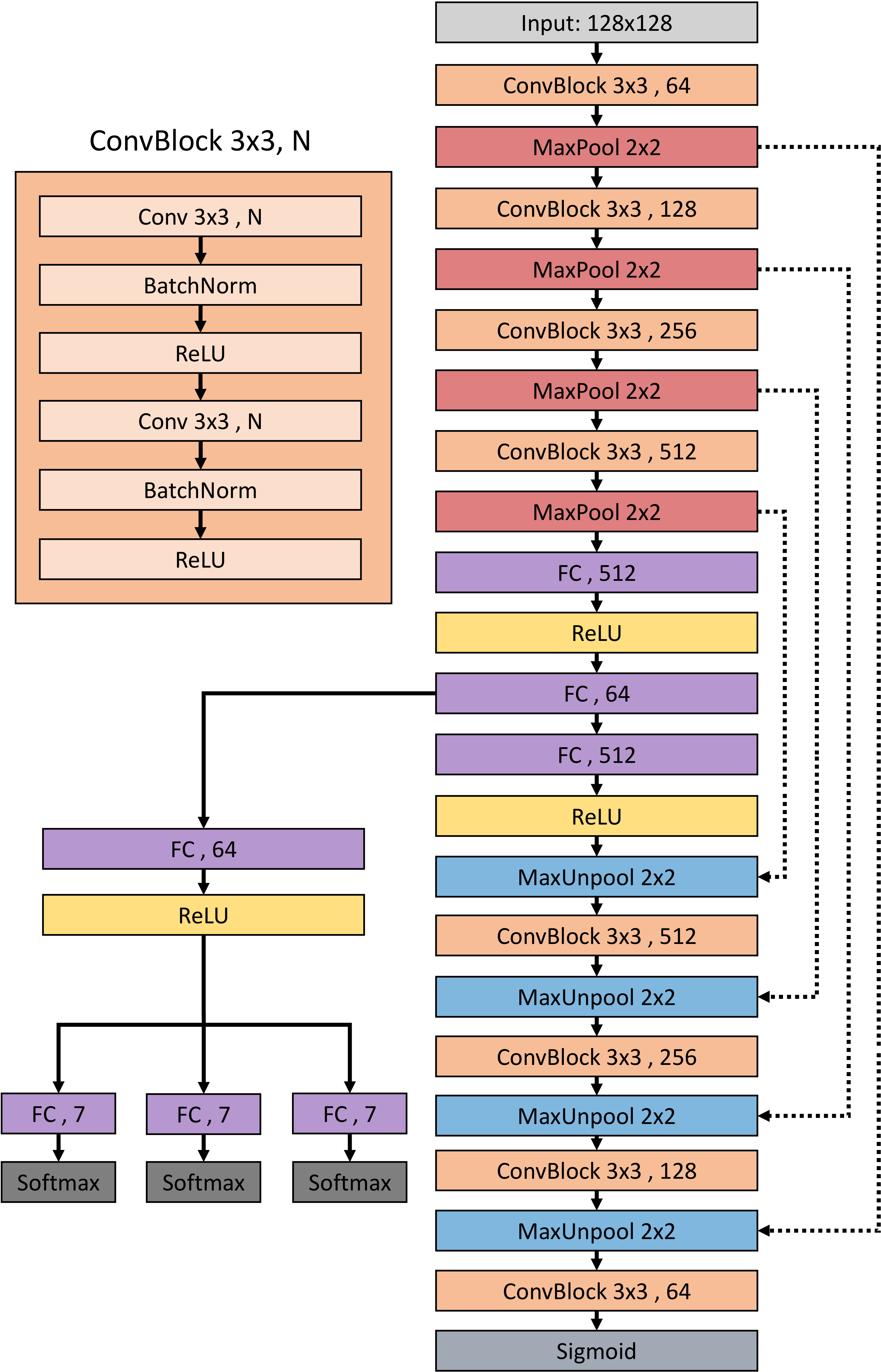}
	\caption{Our model consists of a simple encoder-decoder network with a classifier in the latent space. We use four down-sample and four up-sample blocks: each consisting of two convolutional layers. The number of filters and features are specified for each layer. Max-unpooling (blue) uses the indices from the down-sampling counter-part (red). We use three independent softmax layers for the classification.}
	\label{fig:network}
\end{figure}

Our autoencoder network architecture is inspired by SegNet \cite{badrinarayanan2017segnet}: we keep track of the indices from the max-pooling in the encoder part and use them for the max-unpooling in the decoder part. The network architecture is illustrated in Fig. \ref{fig:network}: All convolutional layers use 3x3 kernels with a padding of 1 and a stride of 1. Each convolutional layer is followed by batch normalization and a ReLU activation function. After each block of two convolutional layers we apply a max-pooling (or max-unpooling) with a kernel size of 2x2 and a stride of 2 and we double (or halve) the amount of filters. The four down-sampling blocks are followed by two fully connected layers (+ intermediate ReLU) where the last one generates the latent vector. The decoder is exactly the reverse of the encoder network with a final sigmoid function. The latent vector is used as input for the decoder network and the classification network. The latter consists of two fully connected layers (+ intermediate ReLU) where the last one outputs three predictions: one for each seat position (left, middle and right). Hence, we need to apply three independent softmax functions to the output of the last layer. This way, we implicitly force the network to learn to use each classifier for a different seat position. Moreover, we can use the whole image as input compared to using an individual image for each seat position \cite{DiasDaCruz2020SVIRO}. Thus, passengers leaning over to the neighbouring seat can be easier classified. Additionally, the middle prediction can be dropped in case of a two-seated car. We will also report results for the same architecture with nearest neighbour up-sampling instead of max-unpooling.  

We define the cost function $\mathcal{L}(x, \hat{x}; \theta, \phi, \omega)$ as a combination of the reconstruction loss for the whole image and the classification loss for each seat position:
\begin{equation}
\begin{split}
	 \mathcal{L}(x, \hat{x}; \theta, \phi, \omega) =  \mathrm{r}&\left(x, \mathrm{h}_{\theta}(\mathrm{g}_{\phi}(\hat{x}))\right) - \\ &\gamma \sum_{i=0}^{2} p_i(x) \log\left(\mathrm{c}_{\omega_i}(\mathrm{g}_{\phi}(\hat{x}))\right) ,
\label{eq:loss}
\end{split}
\end{equation}
where $\mathrm{g}_{\phi}$ is the encoder, $\mathrm{h}_{\theta}$ the decoder, $\mathrm{c}_{\omega_i}$ the classifier for seat $i$ where $i$ corresponds to the left, right and middle seat position, $p_i(x)$ is the true probability distribution for $x$ for seat $i$, and $\gamma$ is a hyperparameter to weight the classification loss.  The reconstruction loss $\mathrm{r(\cdot, \cdot)}$ is computed between the clean input image $x$ (Fig. \ref{fig:augmentation}.a) and the reconstruction (Fig. \ref{fig:augmentation}.c) of the augmented image $\hat{x}$ (Fig. \ref{fig:augmentation}.b). In this work, we consider for the reconstruction loss the mean squared error (MSE): $\mathrm{r}(a,b)=\norm{a-b}_2^2$, the structural similarity index (SSIM) \cite{bergmann2018improving}: $\mathrm{r}(a,b)=1-\mathrm{SSIM}(a,b)$, the multi-scale structural similarity index (MS-SSIM) \cite{wang2003multiscale}: $\mathrm{r}(a,b)=1-\mathrm{MSSSIM}(a,b)$ and the perceptual (PC) loss \cite{hou2017deep}: sum of MSE errors after each pre-trained VGG-16 block. 

%%%%%%%%%%%%%%%%%%%%%%%%%%%%%%%%%%%%%%%%%%%%%%%%%%%%%%%%%%%%%%%%%%%%%%%%%%%%%%%%

\section{Experiments}
We will present a comparison between a representative selection of classification models. We start by establishing a baseline regarding the transferability between different vehicle interiors and their generalization to new class instances. These results will be compared against our autoencoder approach for which we will conduct additional investigations to highlight some of the advantages compared to classification models: the autoencoder is able to transform sceneries from unknown vehicles back to the vehicle it was trained on. The latter is corroborated by a qualitative and quantitative evaluation on real infrared images to show the applicability to a real application, even when trained on synthetic data. 

% ----------------------------------------------------------------------------------
% ----------------------------------------------------------------------------------
% ----------------------------------------------------------------------------------

\subsection{Training Data}
Our experiments were conducted on SVIRO \cite{DiasDaCruz2020SVIRO}, a synthetic dataset for sceneries in the passenger compartment of ten different vehicles. We limited our training on the eight vehicles with three seat positions, but the evaluation will be performed on all ten vehicles such that the models also need to generalize to the two-seated cars. For the later, the model output for the middle seat will be discarded. Each vehicle consists of a train (2500 images) and a test (500 images) split: SVIRO uses different objects (but identical ones across the vehicles) for the train and test split to evaluate the generalization to unknown vehicle interiors either using known or unknown class instances. When referring in the following to training images from unknown vehicles, those images had not been used during training, but they contain class instances of seen objects. SVIRO contains seven different classes: empty seat, occupied and empty infant seat, occupied and empty child seat, adult passenger and everyday object (e.g. bags). We conducted our experiments on the grayscale images (simplified infrared imitations) which helps to become less susceptible to changing illumination \cite{Da_Cruz_2021_WACV}. 

% ----------------------------------------------------------------------------------
% ----------------------------------------------------------------------------------
% ----------------------------------------------------------------------------------\\

\subsection{Training Details}
All models were trained using the following data augmentations: we applied a random horizontal flip (labels are flipped as well) and randomly used transformations from the imgaug 0.4.0 library \cite{imgaug} (PerspectiveTransform, Emboss, Invert, SigmoidContrast, AllChannelsCLAHE and AdditiveLaplaceNoise). The perspective transformation has the largest impact on improving the generalization between the vehicles. Images were center-cropped to 640x640 and then resized to 128x128 for the autoencoder and 224x224 for the classification models. For the latter, the grayscale images were repeated across the channel dimension to form a 3-channel image. The training dataset was partitioned randomly according to a 80:20 split for training and evaluation (the evaluation data was augmented for better transferability assessment), where the latter was used to perform early stopping with respect to classification accuracy. The models were trained for 1000 epochs with a batch size of 64. We used PyTorch 1.4.0 and TorchVision 0.5.0 for all our experiments and PyTorch MS-SSIM \cite{Gongfan2019} for some reconstruction losses. The random seeds of all libraries were fixed for all experiments. Even though the classes are imbalanced \cite{DiasDaCruz2020SVIRO}, we did not consider a weighted loss or imbalanced sampling, because some models are capable of achieving a good performance on the training objects in unknown vehicles. Since SVIRO contains seven classes, all the classification networks output 21 values which corresponds to three predictions: one for each seat position (left, middle and right). The classification loss is computed by adding up three cross-entropy losses (one for each seat position) as formulated in (\ref{eq:loss}). 

% ----------------------------------------------------------------------------------
% ----------------------------------------------------------------------------------
% ----------------------------------------------------------------------------------\\

\subsection{Representative classification models}
\label{section:classification}
We trained six classification models (DenseNet-121, MobileNet V2, ResNet-18, ResNet-50, SqueezeNet V1.1 and VGG-16) as implemented and pre-trained in TorchVision 0.5.0 from scratch or fine-tuned all layers. We replaced the last fully-connected layer with an output similar to the autoencoder classification network  explained in Section \ref{section:method}. Fine-tuning the classification layer or last-block only behaves similarly as shown in the baseline evaluation \cite{DiasDaCruz2020SVIRO}. We used the Adam optimizer with a learning rate of 0.0001. While we used weight decay for the autoencoder training, we decided to not use it for the classification models because of a conducted hyperparameter search. The training images were augmented using the same transformations as for the autoencoder approach. For each classification model and each training method (scratch and fine-tuned), we trained an individual model on the training images of each vehicle and then evaluated it on the training and test images of all the other nine unknown vehicles. The results are summarized in Table \ref{table:overview}. A detailed performance comparison for models trained from scratch on the Tesla vehicle and evaluated on the training images of all unknown vehicles is reported in Table \ref{table:compare_classif}. The models' performances vary a lot across the different vehicle interiors: higher accuracy performance on ImageNet does not guarantee a better inter-vehicle transferability (e.g. DenseNet compared to MobileNet). Moreover, a performance evaluation on one vehicle does not necessarily transfer to a similar performance on other vehicles (e.g. ResNet-50 vs. MobileNet on the A-Class compared to most other vehicles). Hence, it is difficult to assess in advance how well a model might perform in a new vehicle, even if only class instances seen during training are considered. Moreover, evaluating a model's performance on a subset of vehicles does not guarantee a similar behaviour in a different car. Table \ref{table:compare_classif} provides a comparison between training from scratch and fine-tuning classification models against the autoencoder approach. The fine-tuned models achieve overall a higher accuracy, however, the performance still fluctuates between different vehicle interiors. The performance on the Tiguan is worse compared to all other vehicles, because of the unique pattern and color difference in the texture of the rear seats: e.g. see last row of Fig. \ref{fig:transfer}.
\begin{table*}
	\caption{Comparison of the accuracies (in percentage) across different vehicles. The classification models were trained from \textit{\textbf{scratch}} (S) or \textit{\textbf{fine-tuned}} (F) and the autoencoders with (AE) and without (AEW) max-unpooling were trained from scratch only with different reconstruction losses. The models were trained on the augmented training images of the \textbf{\textit{Tesla}} vehicle and tested on the \textbf{\textit{training}} images of all vehicles not seen during training.}
	\begin{center}
        \setlength{\tabcolsep}{3.5pt}
		\begin{tabular}{|c|cc|cc|cc|cc|cc|cc||cc|cc|cc|cc|}
			\cline{2-21}
			\multicolumn{1}{c|}{} & \multicolumn{2}{c|}{VGG-16} & \multicolumn{2}{c|}{DenseNet-121} & \multicolumn{2}{c|}{MobileNet} & \multicolumn{2}{c|}{ResNet-50} & \multicolumn{2}{c|}{ResNet-18} & \multicolumn{2}{c||}{SqueezeNet} & \multicolumn{2}{c|}{SSIM} & \multicolumn{2}{c|}{MS-SSIM} & \multicolumn{2}{c|}{PC} & \multicolumn{2}{c|}{MSE} \tstrut \\
			\hline
			Tested on & F & S & F & S & F & S & F & S & F & S & F & S & AE & AEW & AE & AEW & AE & AEW & AE & AEW\tstrut \\
			\hline
			A-Class & 93.6 & 73.0 & 81.3 & 65.1 & 80.1 & 82.6 & 76.0 & 75.8 & 82.1 & 61.2 & 76.7 & 69.0 & 78.4 & 81.6 & \underline{\textbf{84.0}} & 73.9 & 80.8 & 81.1 & 78.8 & 81.7 \tstrut \\
			\hline
			Escape & 93.6 & \underline{\textbf{88.8}} & 90.2 & 86.2 & 86.1 & 76.2 & 89.0 & 81.8 & 88.4 & 78.5 & 84.8 & 72.8 & 85.5 & 86.0 & 86.1 & 82.7 & 86.8 & 80.7 & 86.4 & 81.3 \tstrut \\
			\hline
			Hilux & 92.5 & 69.9 & 91.4 & 63.7 & 93.3 & 71.8 & 84.8 & 75.8 & 79.0 & 70.1 & 69.2 & 56.6 & 82.5 & 85.9 & \underline{\textbf{87.9}} & 82.1 & 82.3 & 83.9 & 82.0 & 78.1 \tstrut \\
			\hline
			Lexus & 95.3 & \underline{\textbf{87.2}} & 93.4 & 71.1 & 94.9 & 80.4 & 91.1 & 72.8 & 89.3 & 70.1 & 76.8 & 58.4 & 76.0 & 85.0 & 86.8 & 84.1 & 87.0 & 82.5 & 80.6 & 83.3 \tstrut \\
			\hline
			Tiguan & 78.8 & 59.4 & 83.2 & 69.5 & 81.4 & 65.4 & 79.2 & 63.3 & 77.7 & 63.0 & 85.5 & 69.8 & \underline{\textbf{70.9}} & 65.7 & 65.7 & 58.3 & 63.1 & 61.1 & 60.0 & 61.1 \tstrut \\
			\hline
			Tucson & 98.6 & 89.7 & 86.4 & 84.8 & 94.5 & 77.7 & 93.5 & 86.0 & 86.0 & 84.2 & 86.5 & 75.6 & 93.5 & 92.0 & \underline{\textbf{94.7}} & 89.8 & 92.9 & 91.3 & 92.5 & 93.0 \tstrut \\
			\hline
			X5 & 96.5 & 83.4 & 96.7 & 84.1 & 98.0 & 78.1 & 98.0 & 89.8 & 98.3 & 85.6 & 93.6 & 81.6 & \underline{\textbf{90.2}} & 88.7 & 89.6 & 88.5 & 84.5 & 83.4 & 84.6 & 81.8 \tstrut \\
			\hline
			i3 & 99.1 & 80.8 & 99.4 & 91.0 & 98.6 & 77.6 & 96.8 & 89.3 & 98.6 & \underline{\textbf{94.4}} & 95.2 & 80.1 & 87.2 & 90.0 & 93.7 & 88.4 & 94.2 & 90.3 & 93.9 & 88.8 \tstrut \\
			\hline
			Zoe & 98.2 & 72.9 & 77.4 & 70.2 & 96.2 & 80.4 & 83.5 & 86.6 & 77.1 & 71.2 & 81.8 & 62.4 & 87.1 & 91.5 & 91.0 & 84.4 & 90.8 & 89.6 & \underline{\textbf{92.4}} & 91.6 \tstrut \\
			\hline
			\hline
			Mean & 94.0 & 78.3 & 88.8 & 76.2 & 91.5 & 76.7 & 88.0 & 80.1 & 86.3 & 75.4 & 83.3 & 69.6 & 83.5 & 85.1 & \underline{\textbf{86.6}} & 81.4 & 84.7 & 82.7 & 83.5 & 82.3 \tstrut \\
			Std & 5.8 & 9.6 & 6.9 & 9.7 & 6.7 & 4.9 & 7.2 & 8.4 & 7.7 & 10.4 & 7.8 & 8.5 & 6.8 & 7.6 & 8.1 & 9.3 & 8.8 & 8.5 & 9.8 & 8.9 \tstrut \\
			\hline
		\end{tabular}
	\end{center}
	\label{table:compare_classif}
\end{table*}

% ----------------------------------------------------------------------------------
% ----------------------------------------------------------------------------------
% ----------------------------------------------------------------------------------\\

\subsection{Proposed autoencoder classification}
\label{section:autoencoder}
We will present results on the autoencoder architecture introduced in Section \ref{section:method} and compare them against the models obtained in Section \ref{section:classification}. All models use a latent space of dimension 64, the AdamW optimizer with a learning rate of 0.0001 and a weight decay of 0.01. We used $\gamma=75$ in (\ref{eq:loss}) to weight the classification loss with respect to the MSE reconstruction loss according to a conducted hyperparameter search. All other reconstruction losses used $\gamma=1$.

We trained an individual autoencoder on each vehicle and for each reconstruction cost function (MSE, SSIM, MS-SSIM and perceptual loss). If not stated otherwise, we used max-unpooling for the up-sampling. The different autoencoder models are compared in Table \ref{table:compare_classif} against the results of the classification models obtained in Section \ref{section:classification}: the models were trained on the Tesla vehicle and compared on the training images of the nine vehicles not seen during training. Different reconstruction losses influence the transferability of the reconstruction quality, but also the classification accuracy. The MS-SSIM cost function yielded the best accuracy when trained on the augmented training images while the perceptual loss generated the best reconstruction quality: see Section \ref{section:vehicle-domain-transform} for a visual comparison. Comparing the mean performance across all vehicles, all the different autoencoder models outperform the classification models trained from scratch and sometimes even outperforms the fine-tuned models pre-trained on a large amount of data. Autoencoders with nearest neighbour up-sampling perform slightly better with respect to accuracy, but cannot compete with the domain transformation presented in Section \ref{section:vehicle-domain-transform}.

% ----------------------------------------------------------------------------------
% ----------------------------------------------------------------------------------
% ----------------------------------------------------------------------------------

\subsection{SVIRO Benchmark Results}
The SVIRO benchmark reports results on the mean test classification accuracy: a model should be trained in a single vehicle and the mean accuracy on the test images across all vehicles not seen during training is taken as the score. Each of the aforementioned models was trained individually on each vehicle and then evaluated on the nine unknown vehicles: an overview of the performances is reported in Table \ref{table:overview}. We report the mean accuracy on training on the different vehicles and evaluating on all remaining nine vehicles: Autoencoders perform better than training classification models from scratch, but no method outperforms consistently.
\begin{table*}
	\caption{Overview of the SVIRO leaderboard accuracies (in percentage). The models were trained on \textbf{\textit{different}} vehicles and then evaluated on the \textbf{\textit{test}} images of all unknown vehicles. Using the augmented images, the autoencoders with (AE) and without (AEW) max-unpooling and the classification models were trained from \textit{\textbf{scratch}} (S) or \textit{\textbf{fine-tuned}} (F).}
	\begin{center}
        \setlength{\tabcolsep}{3.5pt}
		\begin{tabular}{|c|cc|cc|cc|cc|cc|cc||cc|cc|cc|cc|}
			\cline{2-21}
			\multicolumn{1}{c|}{} & \multicolumn{2}{c|}{VGG-16} & \multicolumn{2}{c|}{DenseNet-121} & \multicolumn{2}{c|}{MobileNet} & \multicolumn{2}{c|}{ResNet-50} & \multicolumn{2}{c|}{ResNet-18} & \multicolumn{2}{c||}{SqueezeNet} & \multicolumn{2}{c|}{SSIM} & \multicolumn{2}{c|}{MS-SSIM} & \multicolumn{2}{c|}{PC} & \multicolumn{2}{c|}{MSE} \tstrut \\
			\hline
			Trained on & F & S & F & S & F & S & F & S & F & S & F & S & AE & AEW & AE & AEW & AE & AEW & AE & AEW\tstrut \\
			\hline
			A-Class & 61.3 & 51.9 & 59.2 & 42.4 & 51.8 & 46.5 & 49.1 & 48.3 & 56.6 & 47.6 & 55.0 & 52.4 & \underline{\textbf{54.8}} & 49.9 & 49.8 & 51.6 & 52.5 & 52.3 & 45.3 & 55.0 \tstrut \\
			\hline
			Escape & 62.6 & 49.3 & 58.1 & 49.4 & 47.9 & 50.9 & 49.7 & 47.1 & 52.3 & 49.4 & 51.1 & 48.8 & 52.0 & 53.5 & 47.2 & 53.7 & \underline{\textbf{56.0}} & 53.1 & 49.5 & 52.9 \tstrut \\
			\hline
			Hilux & 54.1 & 49.7 & 60.9 & 47.6 & 58.3 & 45.1 & 52.5 & 41.3 & 56.2 & 45.6 & 54.5 & 48.4 & 49.1 & \underline{\textbf{51.5}} & 47.8 & 48.5 & 45.2 & 51.2 & 51.1 & 49.9 \tstrut \\
			\hline
			Lexus & 58.1 & 60.6 & 67.4 & 48.2 & 57.0 & 53.6 & 49.1 & 48.5 & 59.5 & 52.8 & 56.4 & 51.2 & 58.2 & \underline{\textbf{61.6}} & 58.4 & 57.6 & 61.2 & 59.7 & 59.0 & 60.6 \tstrut \\
			\hline
			Tesla & 61.8 & 50.6 & 64.5 & 47.3 & 60.6 & 49.4 & 57.4 & 51.9 & 57.2 & 47.8 & 50.2 & 46.1 & 55.9 & 55.7 & 58.0 & 55.5 & 54.3 & \underline{\textbf{58.3}} & 58.2 & 58.2 \tstrut \\
			\hline
			Tiguan & 48.5 & 44.6 & 49.2 & 35.5 & 45.5 & 44.7 & 50.0 & 35.7 & 51.2 & 40.9 & 53.4 & 43.4 & 42.1 & 41.9 & 41.2 & 43.2 & 39.1 & 44.8 & \underline{\textbf{46.8}} & 46.2 \tstrut \\
			\hline
			Tucson & 66.3 & 45.9 & 68.4 & 46.6 & 53.0 & 47.9 & 60.0 & 44.7 & 58.7 & 49.9 & 51.4 & 50.6 & 50.4 & 53.6 & 53.8 & 52.1 & 46.1 & \underline{\textbf{55.1}} & 44.5 & 47.6 \tstrut \\
			\hline
			X5 & 58.3 & 52.0 & 52.0 & 40.0 & 51.2 & 47.4 & 49.2 & 43.1 & 42.4 & 46.8 & 45.7 & 39.2 & 53.3 & \underline{\textbf{59.1}} & 50.0 & 52.0 & 52.9 & 53.9 & 55.0 & 52.9 \tstrut \\
			\hline
			\hline
			Mean & 58.9 & 50.6 & 60.0 & 44.6 & 53.2 & 48.2 & 52.1 & 45.1 & 54.3 & 47.6 & 52.2 & 47.5 & 52.0 & 53.3 & 50.8 & 51.8 & 50.9 & \underline{\textbf{53.6}} & 51.2 & 52.9 \tstrut \\
			Std & 5.2 & 4.5 & 6.4 & 4.5 & 4.9 & 2.8 & 4.0 & 4.7 & 5.2 & 3.3 & 3.2 & 4.1 & 4.6 & 5.6 & 5.4 & 4.1 & 6.6 & 4.3 & 5.3 & 4.7 \tstrut \\
			\hline
		\end{tabular}
	\end{center}
	\label{table:overview}
\end{table*}

% ----------------------------------------------------------------------------------
% ----------------------------------------------------------------------------------
% ----------------------------------------------------------------------------------

\subsection{Ablation Study}
\begin{figure}
 	\centering
  	\includegraphics[width=0.95\linewidth]{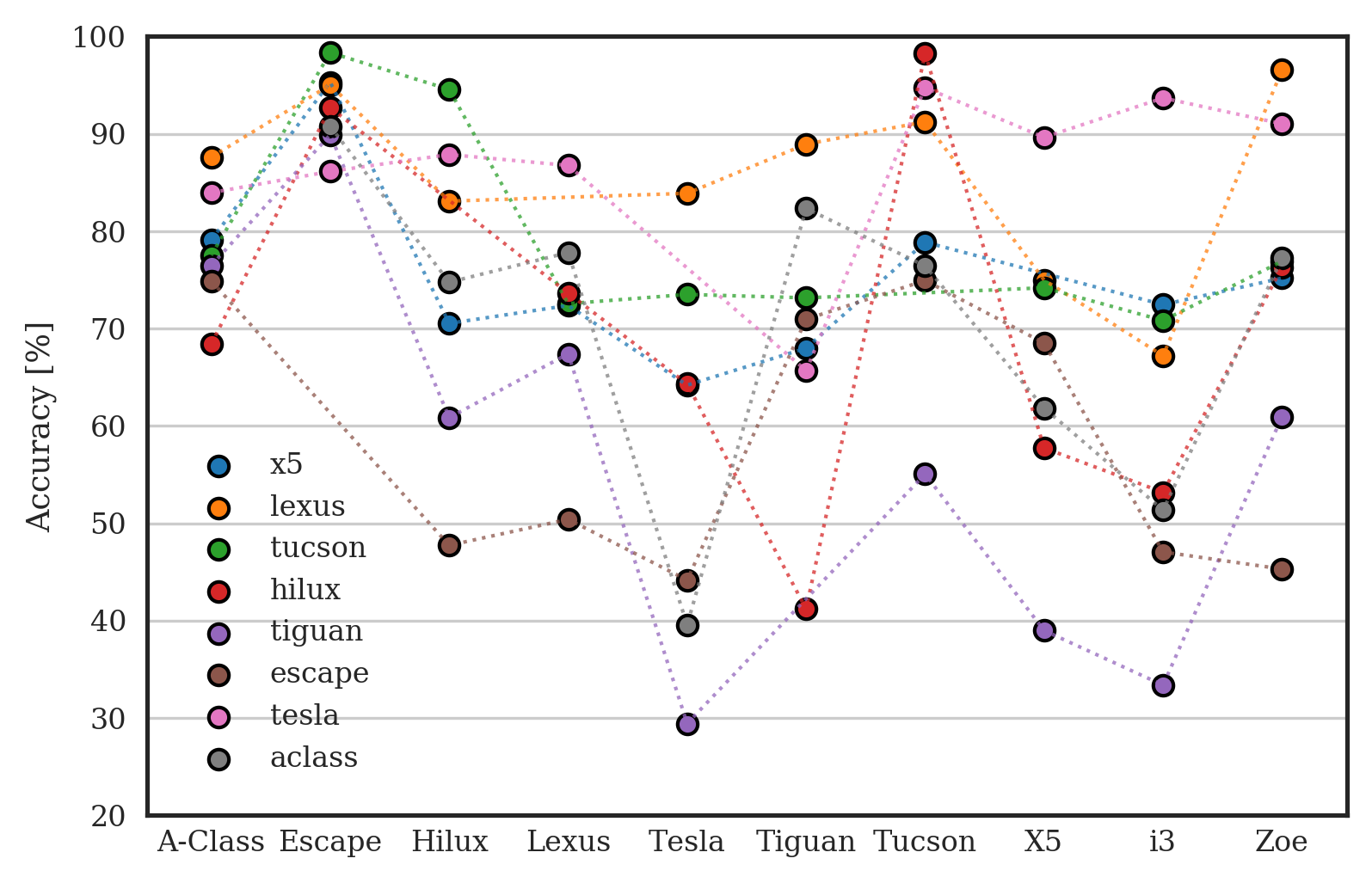}
	\caption{We trained an individual MS-SSIM autoencoder on each of the eight vehicles. The models were evaluated on the training images of the nine vehicles not seen during training. Different colors represent the vehicles each model was trained on. The performances of each model across all unseen vehicles are connected by lines to ease visualization.}
	\label{fig:compare_each_ae}
\end{figure}
\begin{figure}
	\centering
	\includegraphics[width=0.93\linewidth]{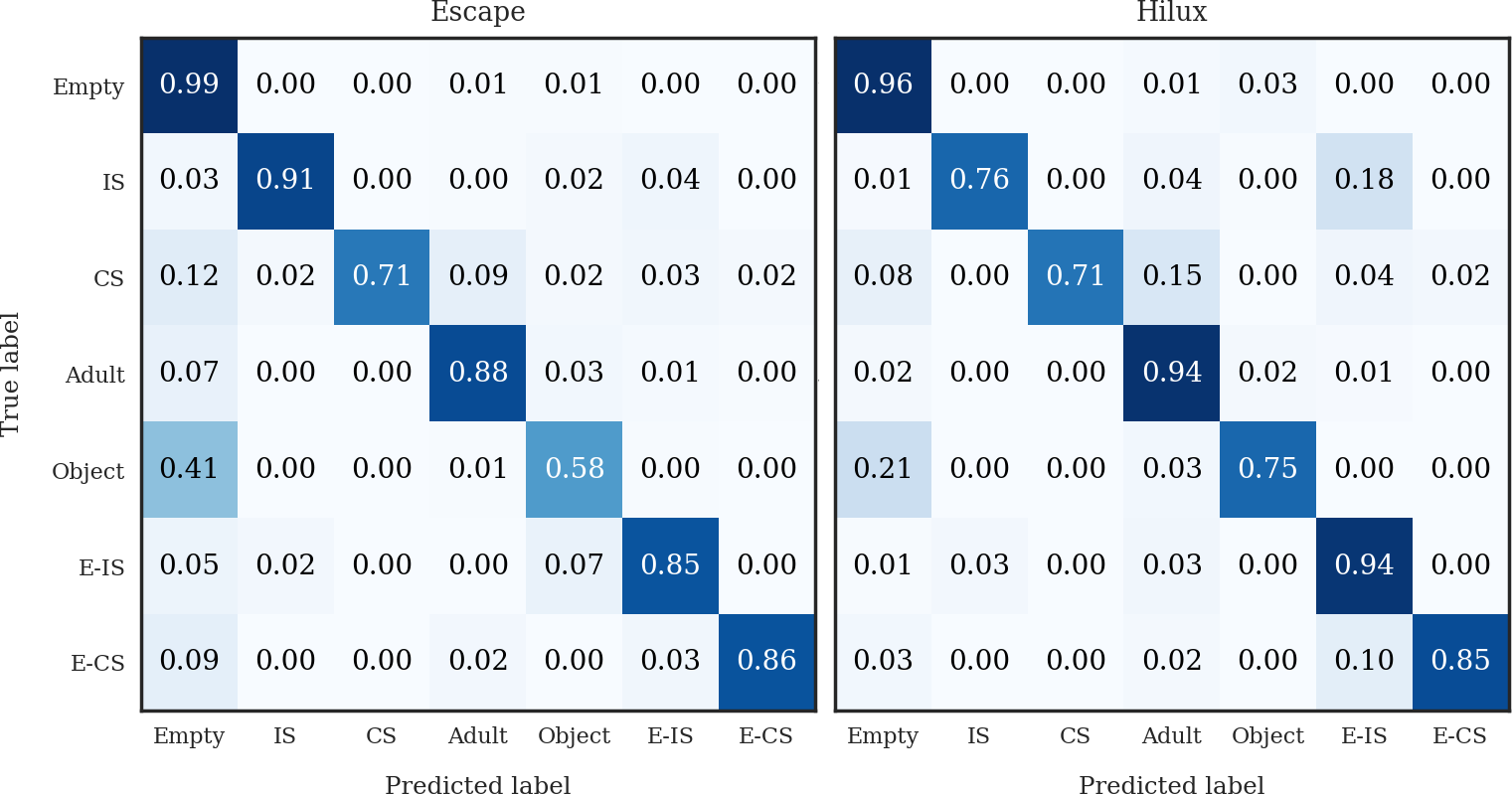}
	\caption{Confusion matrices for the MS-SSIM autoencoder from Fig. \ref{fig:compare_each_ae} trained on the Tesla. Evaluation was done on the training images from the Escape (left) and Hilux (right). Although the model achieves a similar overall accuracy on both vehicles, the mis-classifications are different. Abbreviations: CS = child seat, IS = infant seat and E-* = empty.}
	\label{fig:confusion}
\end{figure}

Apart from comparing the autoencoder based approach to the classification models, we also investigated the method itself in more details. We trained an individual MS-SSIM autoencoder on each of the eight vehicles. In Fig. \ref{fig:compare_each_ae}, the resulting models are compared individually on the training images of each of the nine vehicles not seen during training. Depending on which vehicle the models were trained on, the trend in their overall performance can be quite different and fluctuate a lot: some vehicles are more advantageous while others lead to worse results. In Fig. \ref{fig:confusion}, we compare the confusion matrices from two unknown vehicles with similar mean accuracy by the model trained on the Tesla vehicle: in the Escape the model performs worse on adults and everyday objects and it misclassified many samples as empty while in the Hilux more infant seats are misclassified as unoccupied. Hence, a same model can behave differently (even with similar mean accuracy) on different vehicles such that no guarantees can be provided without additional precautions. 

% ----------------------------------------------------------------------------------
% ----------------------------------------------------------------------------------
% ----------------------------------------------------------------------------------

\subsection{Vehicle Domain Transformation}
\label{section:vehicle-domain-transform}
\begin{figure}
	\centering
	\begin{overpic}[width=0.18\linewidth]{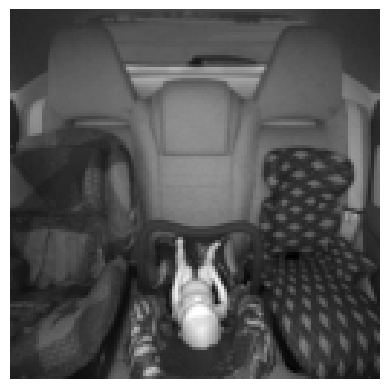}
		\put(17,80){\small\textcolor{white}{A-Class}}
	\end{overpic}
	\begin{overpic}[width=0.18\linewidth]{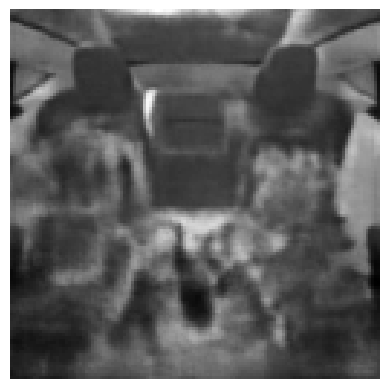}
		\put(27,80){\small\textcolor{white}{MSE}}
	\end{overpic}
	\begin{overpic}[width=0.18\linewidth]{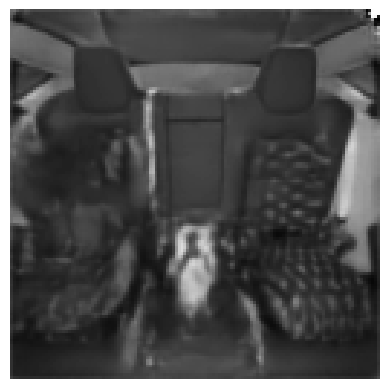}
		\put(26,80){\small\textcolor{white}{SSIM}}
	\end{overpic}
	\begin{overpic}[width=0.18\linewidth]{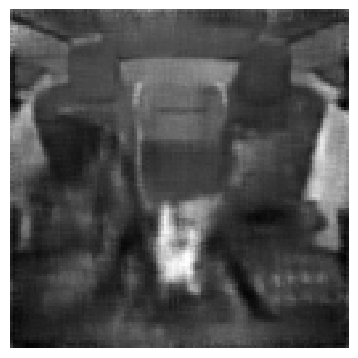}
		\put(9,80){\small\textcolor{white}{MS-SSIM}}
	\end{overpic}
	\begin{overpic}[width=0.18\linewidth]{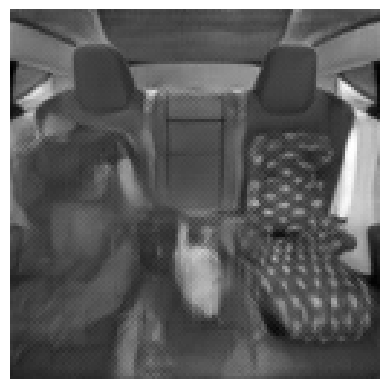}
		\put(7,80){\small\textcolor{white}{Perceptual}}
	\end{overpic}
	\vskip 0.1 \baselineskip
	\begin{overpic}[width=0.19\linewidth]{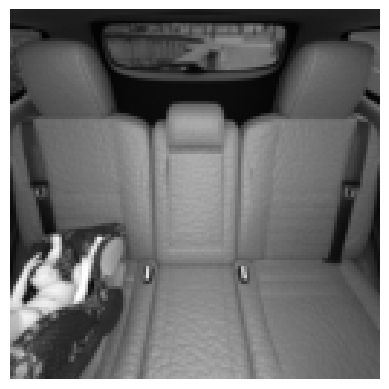}
		\put(23,80){\small\textcolor{white}{Escape}}
	\end{overpic}
	\includegraphics[width=0.18\linewidth]{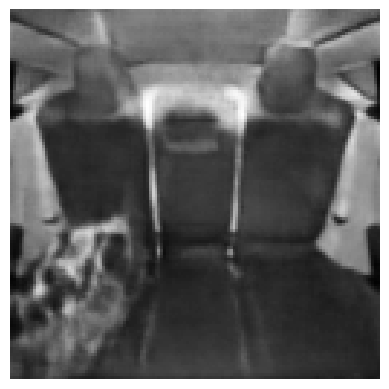}
	\includegraphics[width=0.18\linewidth]{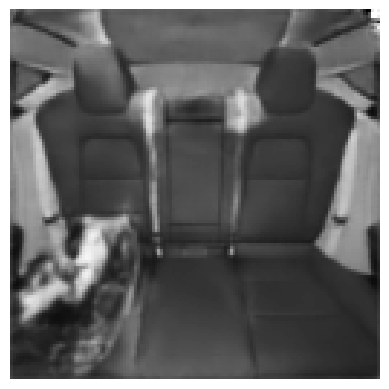}
	\includegraphics[width=0.18\linewidth]{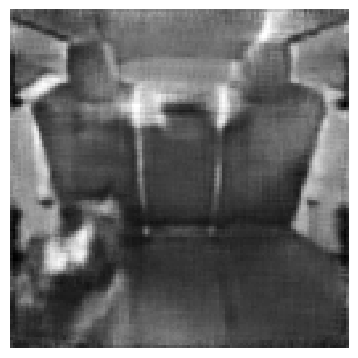}
	\includegraphics[width=0.18\linewidth]{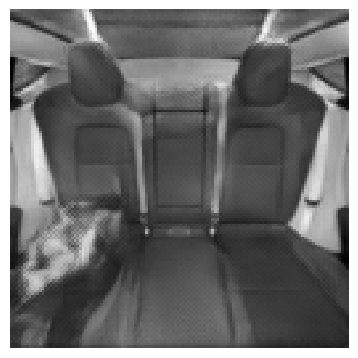}
	\vskip 0.1 \baselineskip
	\begin{overpic}[width=0.18\linewidth]{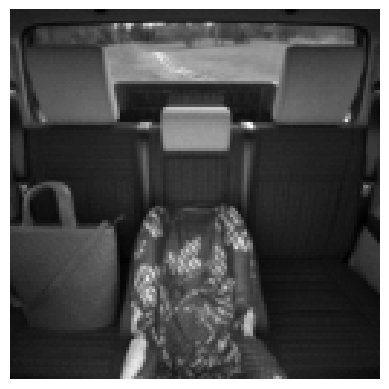}
		\put(28,80){\small\textcolor{white}{Hilux}}
	\end{overpic}
	\includegraphics[width=0.18\linewidth]{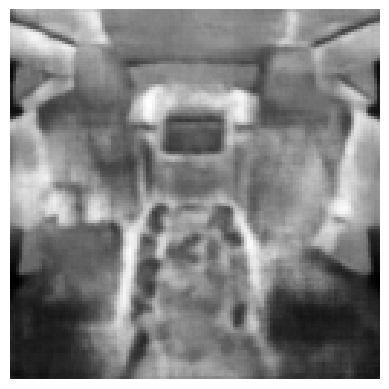}
	\includegraphics[width=0.18\linewidth]{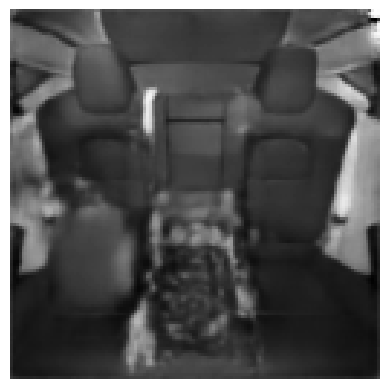}
	\includegraphics[width=0.18\linewidth]{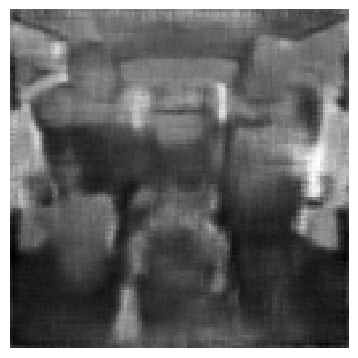}
	\includegraphics[width=0.18\linewidth]{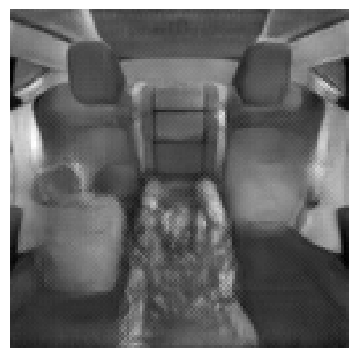}
	\vskip 0.1 \baselineskip
	\begin{overpic}[width=0.18\linewidth]{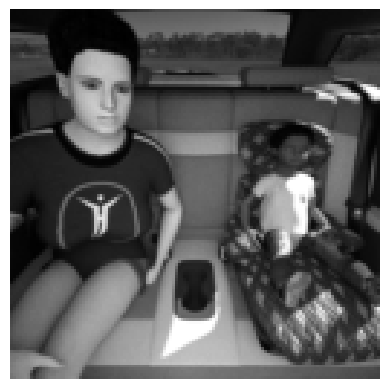}
		\put(44,80){\small\textcolor{white}{i3}}
	\end{overpic}
	\includegraphics[width=0.18\linewidth]{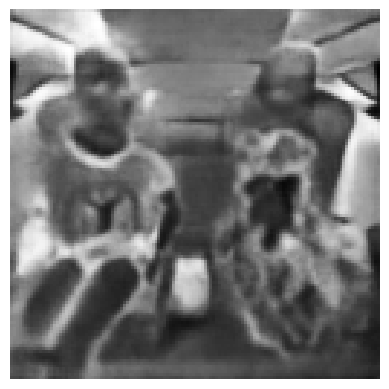}
	\includegraphics[width=0.18\linewidth]{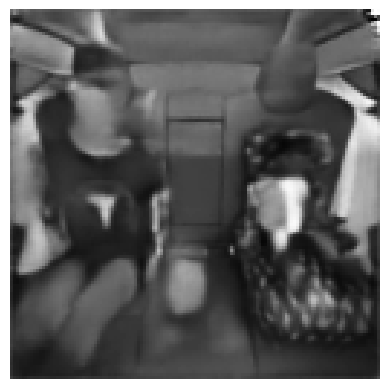}
	\includegraphics[width=0.18\linewidth]{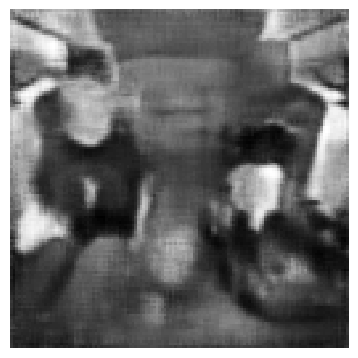}
	\includegraphics[width=0.18\linewidth]{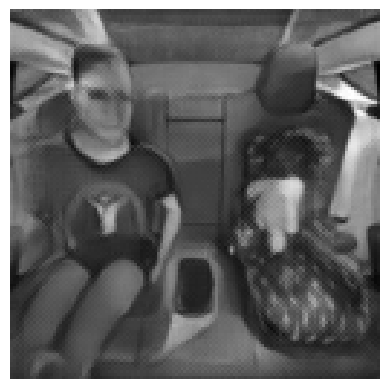}
	\vskip 0.1 \baselineskip
	\begin{overpic}[width=0.18\linewidth]{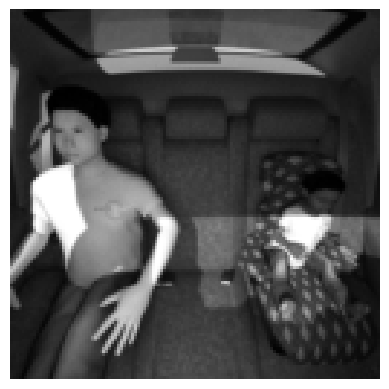}
		\put(26,80){\small\textcolor{white}{Lexus}}
	\end{overpic}
	\includegraphics[width=0.18\linewidth]{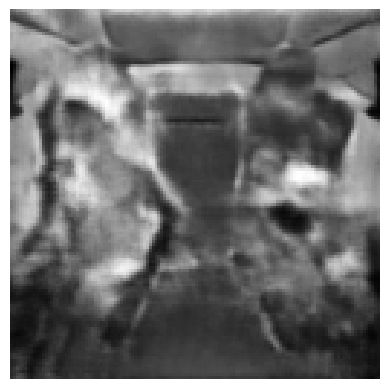}
	\includegraphics[width=0.18\linewidth]{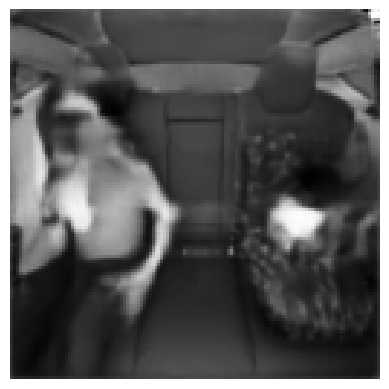}
	\includegraphics[width=0.18\linewidth]{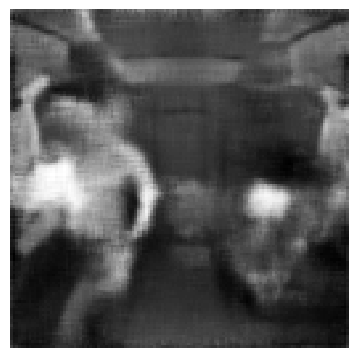}
	\includegraphics[width=0.18\linewidth]{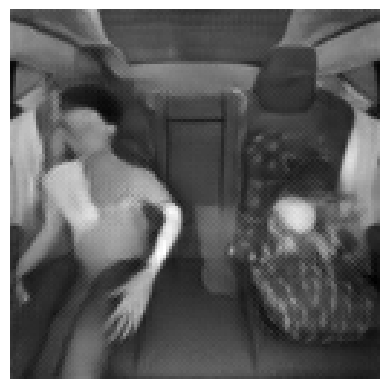}
	\vskip 0.1 \baselineskip
	\begin{overpic}[width=0.18\linewidth]{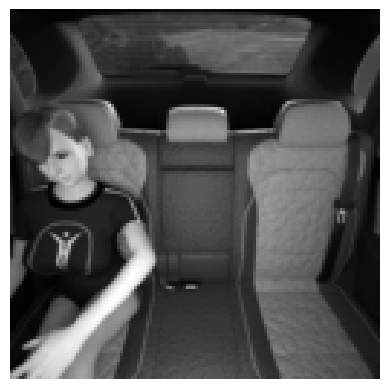}
		\put(21,80){\small\textcolor{white}{Tiguan}}
	\end{overpic}
	\includegraphics[width=0.18\linewidth]{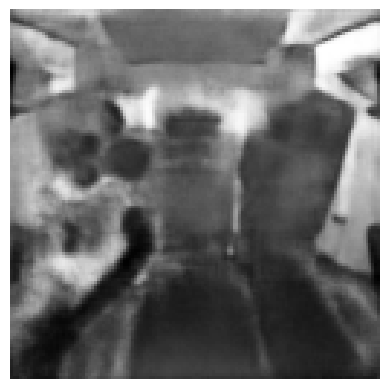}
	\includegraphics[width=0.18\linewidth]{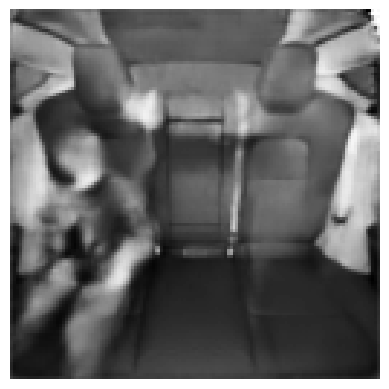}
	\includegraphics[width=0.18\linewidth]{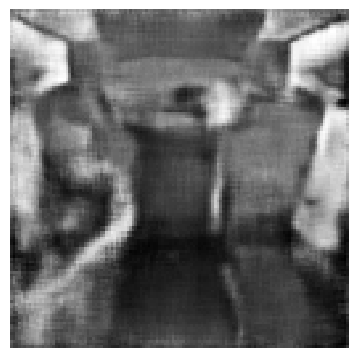}
	\includegraphics[width=0.18\linewidth]{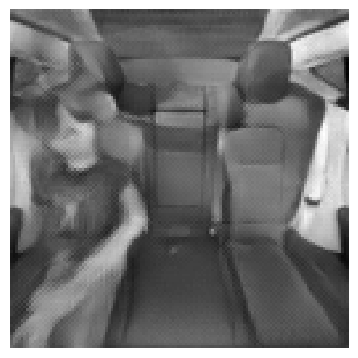}
	\caption{Autoencoders were trained on the Tesla using the de-noising approach. The first column contains input training images from unknown vehicles. The other columns show the corresponding transformations by the autoencoder for different cost functions: MSE, SSIM, MS-SSIM and PC-loss.}
	\label{fig:transfer}
\end{figure}

\begin{figure}
	\centering
	\begin{overpic}[width=0.18\linewidth]{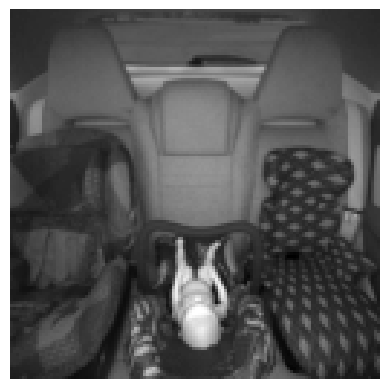}
		\put(28,80){\small\textcolor{white}{Input}}
	\end{overpic}
	\begin{overpic}[width=0.18\linewidth]{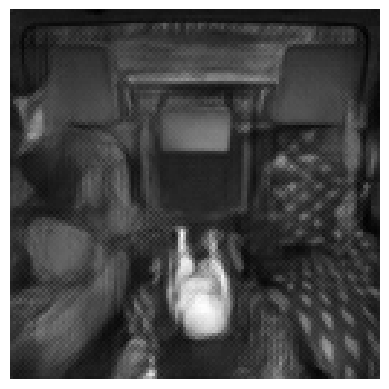}
		\put(29,80){\small\textcolor{white}{Hilux}}
	\end{overpic}
	\begin{overpic}[width=0.18\linewidth]{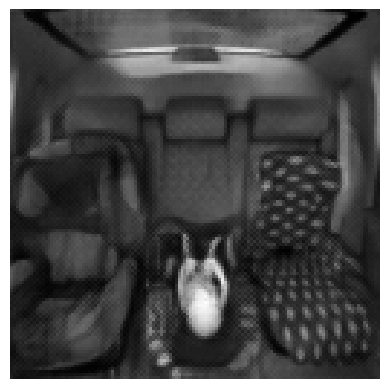}
		\put(27,80){\small\textcolor{white}{Lexus}}
	\end{overpic}
	\begin{overpic}[width=0.18\linewidth]{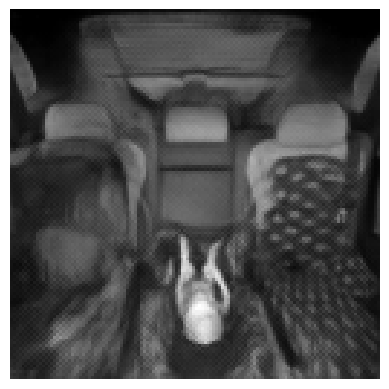}
		\put(24,80){\small\textcolor{white}{Tiguan}}
	\end{overpic}
	\begin{overpic}[width=0.18\linewidth]{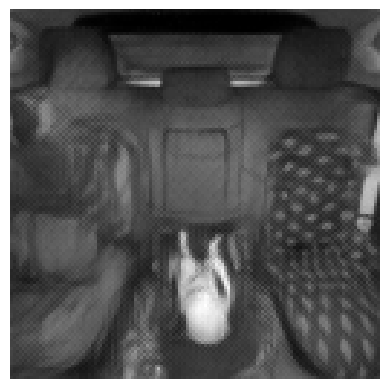}
		\put(22,80){\small\textcolor{white}{Tucson}}
	\end{overpic}
	\vskip 0.1 \baselineskip
	\includegraphics[width=0.18\linewidth]{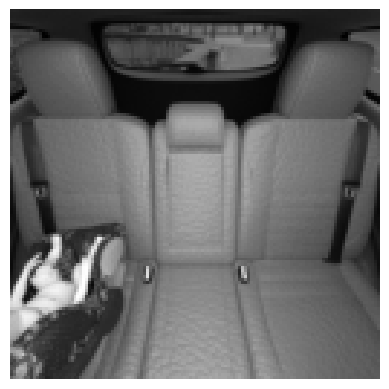}
	\includegraphics[width=0.18\linewidth]{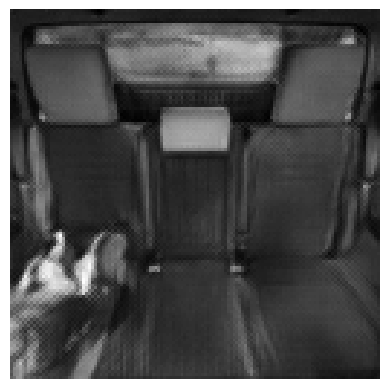}
	\includegraphics[width=0.18\linewidth]{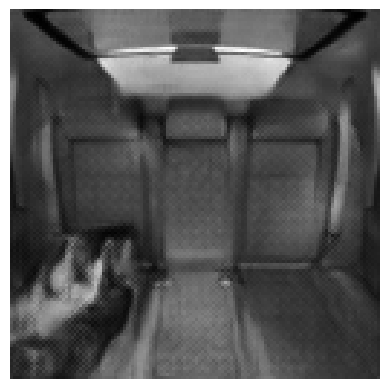}
	\includegraphics[width=0.18\linewidth]{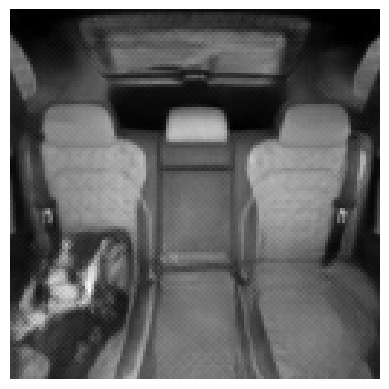}
	\includegraphics[width=0.18\linewidth]{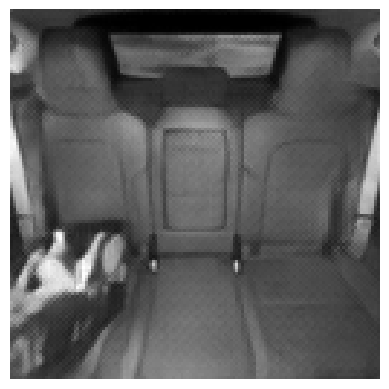}
	\vskip 0.1 \baselineskip
	\includegraphics[width=0.18\linewidth]{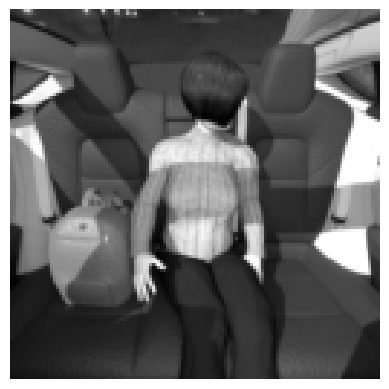}
	\includegraphics[width=0.18\linewidth]{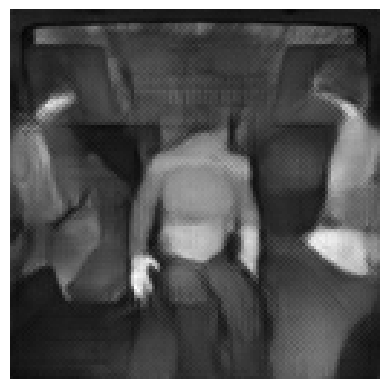}
	\includegraphics[width=0.18\linewidth]{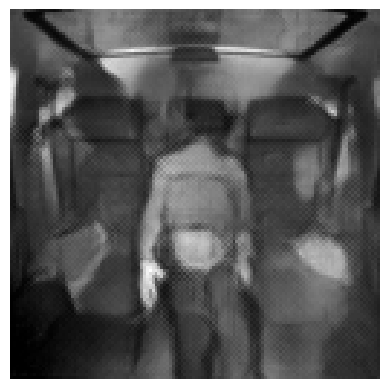}
	\includegraphics[width=0.18\linewidth]{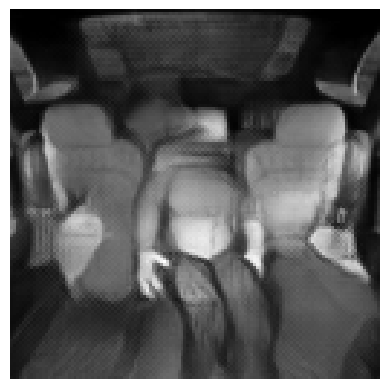}
	\includegraphics[width=0.18\linewidth]{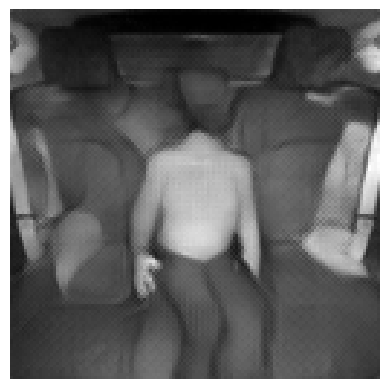}
	\caption{Reconstruction of the same sceneries (first column) by autoencoders trained on different vehicles (remaining columns) using the perceptual loss during training.}
	\label{fig:per-vehicle-recon}
\end{figure}

An additional advantage of the autoencoder approach is the possibility to exploit the disadvantage of our training environment: since all images are from the same vehicle, the sceneries will not undergo many changes (besides the objects on the seat). Hence, by augmenting the training images and using the autoencoder as a de-noiser, the model learns to transform images back to the original environment. This leads to useful properties when images from an unknown vehicle are used as input. In Fig. \ref{fig:transfer} we compare input images from six unknown vehicles against their transformed versions computed by autoencoders from Table \ref{table:compare_classif} trained with different cost functions on images from the Tesla. The autoencoders are able to transform the input images to a scenery in the vehicle the models were trained on by replacing the rear bench and adapting the perspective accordingly. The different cost functions have an influence on the transferability and quality of the transformations. Overall, SSIM and the perceptual loss perform best. The reconstruction of the objects is not perfect, especially humans often appear blurrier, probably because of the higher visual complexity and variability due to different poses. Notice that it is much harder for the image from the Tiguan, because of the unique texture pattern. The reconstructions for the objects need further investigations regarding the delivery of valid and robust features that can be used from a classifier. In Fig. \ref{fig:per-vehicle-recon} we show several examples of the reconstruction of the same sceneries by autoencoders trained on different vehicles using the perceptual loss. For nearest neighbour up-sampling the models have more trouble for the reconstruction in unseen vehicles, because all the information goes through the latent space in contrast to the indices when max-unpooling is used. 

% ----------------------------------------------------------------------------------
% ----------------------------------------------------------------------------------
% ----------------------------------------------------------------------------------

\subsection{Applicability to Real Infrared Images}
\label{section:real}
\begin{figure}
	\centering
	\begin{overpic}[width=0.20\linewidth]{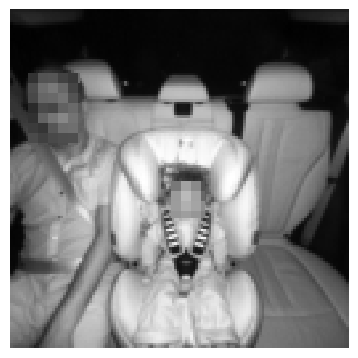}
		\put(31,82){\small\textcolor{white}{Input}}
	\end{overpic}
	\includegraphics[width=0.20\linewidth]{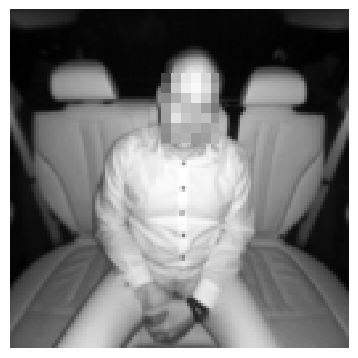}
	\includegraphics[width=0.20\linewidth]{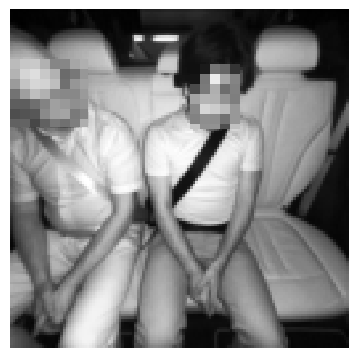}
	\includegraphics[width=0.20\linewidth]{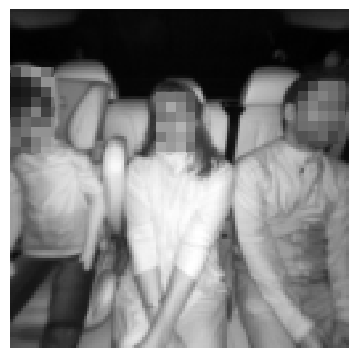}
	\vskip 0.1 \baselineskip
	\begin{overpic}[width=0.20\linewidth]{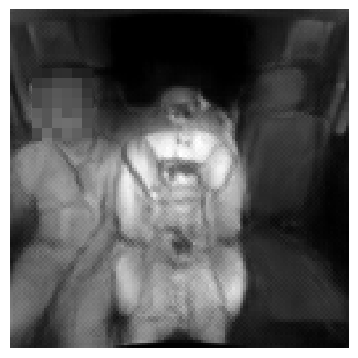}
		\put(11,82){\small\textcolor{white}{Recon - R}}
	\end{overpic}
	\includegraphics[width=0.20\linewidth]{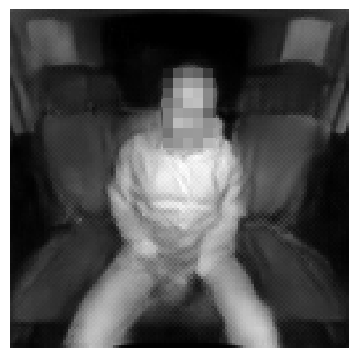}
	\includegraphics[width=0.20\linewidth]{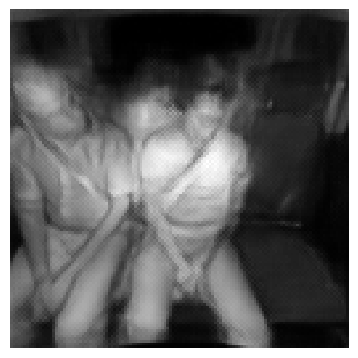}
	\includegraphics[width=0.20\linewidth]{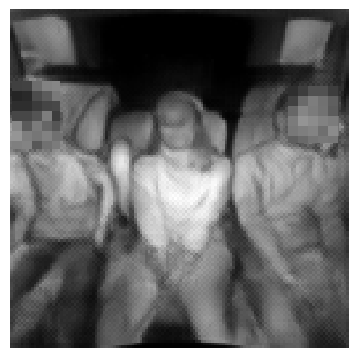}
	\vskip 0.1 \baselineskip
	\begin{overpic}[width=0.20\linewidth]{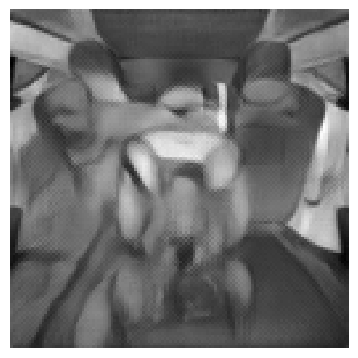}
		\put(11,82){\small\textcolor{white}{Recon - S}}
	\end{overpic}
	\includegraphics[width=0.20\linewidth]{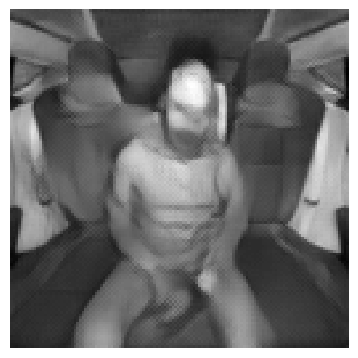}
	\includegraphics[width=0.20\linewidth]{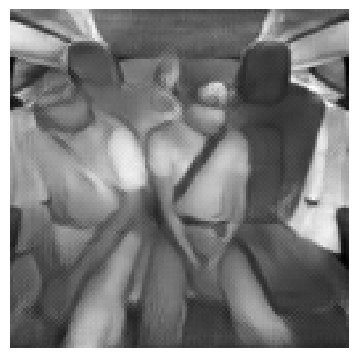}
	\includegraphics[width=0.20\linewidth]{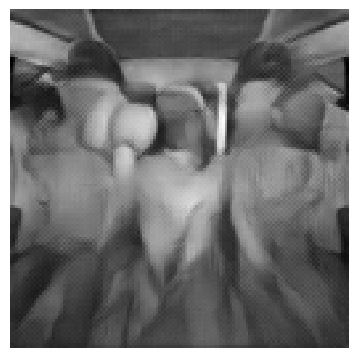}
	\vskip 0.1 \baselineskip
	\begin{overpic}[width=0.20\linewidth]{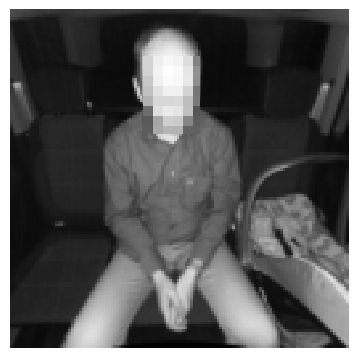}
		\put(31,82){\small\textcolor{white}{Input}}
	\end{overpic}
	\includegraphics[width=0.20\linewidth]{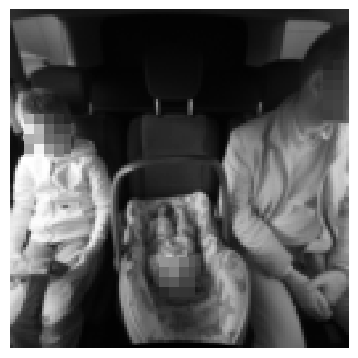}
	\includegraphics[width=0.20\linewidth]{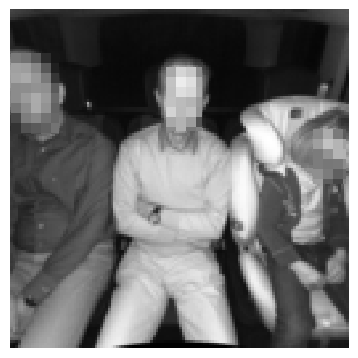}
	\includegraphics[width=0.20\linewidth]{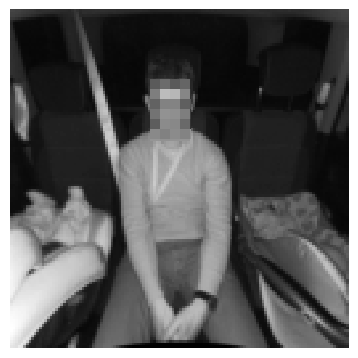}
	\vskip 0.1 \baselineskip
	\begin{overpic}[width=0.20\linewidth]{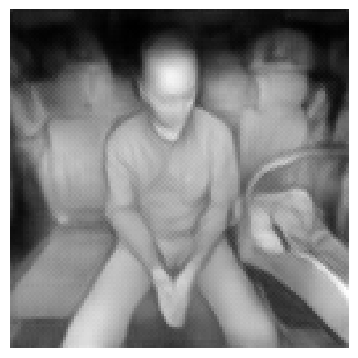}
		\put(11,82){\small\textcolor{white}{Recon - R}}
	\end{overpic}
	\includegraphics[width=0.20\linewidth]{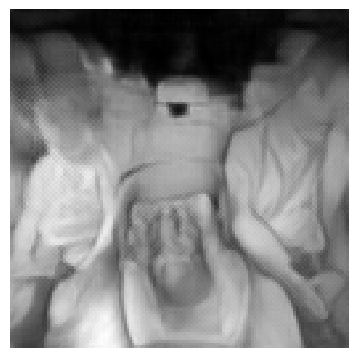}
	\includegraphics[width=0.20\linewidth]{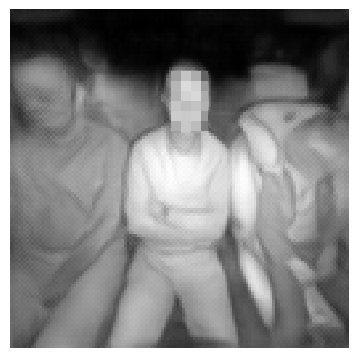}
	\includegraphics[width=0.20\linewidth]{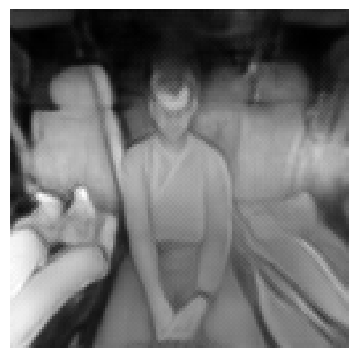}
	\vskip 0.1 \baselineskip
	\begin{overpic}[width=0.20\linewidth]{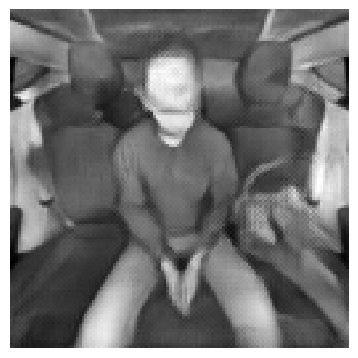}
		\put(11,82){\small\textcolor{white}{Recon - S}}
	\end{overpic}
	\includegraphics[width=0.20\linewidth]{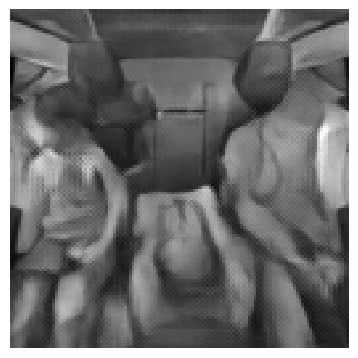}
	\includegraphics[width=0.20\linewidth]{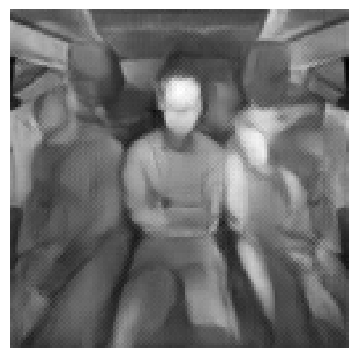}
	\includegraphics[width=0.20\linewidth]{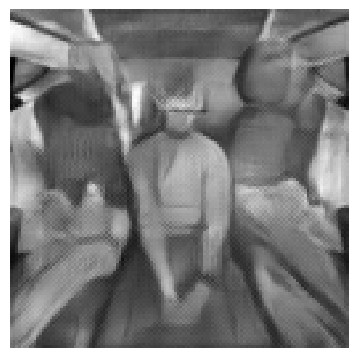}
	\caption{Individual models were trained on \textit{real} images from an X5 (first row) and Sharan (fourth row) and on \textit{synthetic} images from the Tesla using the perceptual loss. The models were then applied to real images from the vehicle not used during training. Recon-R are reconstructions when trained on real images and Recon-S when trained on synthetic ones.}
	\label{fig:real}
\end{figure}

We tested the transferability of the vehicle domain transformation presented in Section \ref{section:vehicle-domain-transform} to real images. To this end, we recorded 3500 sceneries by an active infrared camera system in two vehicle interiors: BMW X5 and VW Sharan. We used the same classes as for the synthetic dataset and individual autoencoders were then trained on both real vehicles and the synthetic Tesla images. We used the same architecture as presented in Section \ref{section:method} and the perceptual loss for training. In Fig. \ref{fig:real}, we report results on the autoencoder transformations for images of the vehicle not used during training. The backgrounds and rear seats from the Sharan are replaced by the ones from the X5 and vice versa. Albeit the results are not as detailed as if trained on real images, the transfer from synthetic to real images is possible and the background is replaced by the synthetic one. Additionally, we trained individual classification models and autoencoders using several reconstruction cost functions in order to compare their generalization accuracy. The results are summarized in Table \ref{table:compare_classif_real} and they are similar as for the synthetic data: our autoencoder approach generalizes better than classification models trained from scratch. Fine-tuned classification models usually perform best, but we think that improvements on the autoencoder will further close the gap and yield additional helpful properties (e.g. disentanglement). 

\begin{table*}%[ht]
	\caption{Classification models were trained from \textbf{\textit{scratch}} (S) or \textit{\textbf{fine-tuned}} (F) and autoencoders with (AE) and without (AEW) max-unpooling were trained using different reconstruction losses. The models were trained on the augmented real images from the Sharan and X5 respectively and evaluated [\%] on the images from the vehicle not seen during training.}
	\begin{center}
        \setlength{\tabcolsep}{3.5pt}
		\begin{tabular}{|c|cc|cc|cc|cc|cc|cc||cc|cc|cc|cc|}
			\cline{2-21}
			\multicolumn{1}{c}{} & \multicolumn{2}{|c|}{VGG-16} & \multicolumn{2}{c|}{DenseNet-121} & \multicolumn{2}{c|}{MobileNet} & \multicolumn{2}{c|}{ResNet-50} & \multicolumn{2}{c|}{ResNet-18} & \multicolumn{2}{c||}{SqueezeNet} & \multicolumn{2}{c|}{SSIM} & \multicolumn{2}{c|}{MS-SSIM} & \multicolumn{2}{c|}{PC} & \multicolumn{2}{c|}{MSE} \tstrut \\
			\hline
			Tested on & F & S & F & S & F & S & F & S & F & S & F & S & AE & AEW & AE & AEW & AE & AEW & AE & AEW\tstrut \\
			\hline
			Sharan & 90.2 & 77.5 & 95.4 & 68.1 & 91.9 & 70.1 & 91.2 & 65.6 & 87.7 & 68.7 & 79.9 & 70.6 & \underline{\textbf{84.9}} & 75.3 & 76.8 & 78.3 & 75.7 & 73.9 & 77.2 & 71.6 \tstrut \\
			\hline
			X5 & 87.9 & 74.7 & 85.2 & 68.2 & 89.3 & 60.1 & 93.0 & 66.8 & 93.0 & 74.4 & 84.3 & 57.1 & 76.2 & 81.7 & 82.1 & 78.1 & 81.8 & \underline{\textbf{88.4}} & 76.9 & 72.5 \tstrut \\
			\hline
			\hline
			Mean & 89.1 & 76.1 & 90.3 & 68.2 & 90.6 & 65.1 & 92.1 & 66.2 & 90.4 & 71.6 & 82.1 & 63.8 & \underline{\textbf{80.6}} & 78.5 & 79.4 & 78.2 & 78.8 & \underline{\textbf{81.2}} & 77.1 & 72.0 \tstrut \\
			\hline
		\end{tabular}
	\end{center}
	\label{table:compare_classif_real}
\end{table*}

%%%%%%%%%%%%%%%%%%%%%%%%%%%%%%%%%%%%%%%%%%%%%%%%%%%%%%%%%%%%%%%%%%%%%%%%%%%%%%%%

\section{CONCLUSION and FUTURE WORK}
We introduced the challenge of training in a single vehicle interior and improving generalization to unknown vehicles and class instances. Our results showed that commonly used classifiers do not behave reliably across different vehicle interiors and our introduced autoencoder approach outperforms classification models trained from scratch. This is important when pre-trained models cannot be used for commercialized applications due to licensing constraints. Although the applicability to real images has been shown, the generalization to new class instances needs to be improved and the transfer to new vehicles needs to be robustified to be applicable to safety critical applications. None of the investigated classification and autoencoder methods can guarantee a similar behaviour on potentially new vehicles, even when multiple vehicles are available to test the models' behaviours during the design process. The models' predictions should be accompanied with uncertainty estimations to quantify the model's self-assessment with respect to its capability to generalize to new vehicle interiors or new sceneries. We believe that improvements on autoencoders will outperform classification models further for our problem formulation. Additional constraints and latent space properties like disentanglement can be applied and might benefit our problem statement as well. 

%%%%%%%%%%%%%%%%%%%%%%%%%%%%%%%%%%%%%%%%%%%%%%%%%%%%%%%%%%%%%%%%%%%%%%%%%%%%%%%%

\section*{ACKNOWLEDGMENT}

The first author is supported by the Luxembourg National Research Fund (13043281). This work was supported by the Luxembourg Ministry of the Economy (CVN 18/18/RED).

%%%%%%%%%%%%%%%%%%%%%%%%%%%%%%%%%%%%%%%%%%%%%%%%%%%%%%%%%%%%%%%%%%%%%%%%%%%%%%%%

\bibliographystyle{IEEEtran}
\bibliography{IEEEabrv, bib}

\begin{thebibliography}{10}
\providecommand{\url}[1]{#1}
\csname url@rmstyle\endcsname
\providecommand{\newblock}{\relax}
\providecommand{\bibinfo}[2]{#2}
\providecommand\BIBentrySTDinterwordspacing{\spaceskip=0pt\relax}
\providecommand\BIBentryALTinterwordstretchfactor{4}
\providecommand\BIBentryALTinterwordspacing{\spaceskip=\fontdimen2\font plus
\BIBentryALTinterwordstretchfactor\fontdimen3\font minus
  \fontdimen4\font\relax}
\providecommand\BIBforeignlanguage[2]{{%
\expandafter\ifx\csname l@#1\endcsname\relax
\typeout{** WARNING: IEEEtran.bst: No hyphenation pattern has been}%
\typeout{** loaded for the language `#1'. Using the pattern for}%
\typeout{** the default language instead.}%
\else
\language=\csname l@#1\endcsname
\fi
#2}}

\bibitem{airbag}
M.~E. Farmer and A.~K. Jain, ``Occupant classification system for automotive
  airbag suppression,'' in \emph{Conference on Computer Vision and Pattern
  Recognition (CVPR)}, 2003.

\bibitem{perrett2016cost}
T.~Perrett and M.~Mirmehdi, ``Cost-based feature transfer for vehicle occupant
  classification,'' in \emph{Asian Conference on Computer Vision (ACCV)}, 2016.

\bibitem{tian2018eliminating}
M.~Tian, S.~Yi, H.~Li, S.~Li, X.~Zhang, J.~Shi, J.~Yan, and X.~Wang,
  ``Eliminating background-bias for robust person re-identification,'' in
  \emph{Conference on Computer Vision and Pattern Recognition (CVPR)}, 2018.

\bibitem{da2019theoretical}
S.~Dias Da~Cruz, H.-P. Beise, U.~Schr{\"o}der, and U.~Karahasanovic, ``A
  theoretical investigation of the detection of vital signs in presence of car
  vibrations and radar-based passenger classification,'' \emph{Transactions on
  Vehicular Technology (TVT)}, 2019.

\bibitem{pan2009survey}
S.~J. Pan and Q.~Yang, ``A survey on transfer learning,'' \emph{IEEE
  Transactions on knowledge and data engineering}, vol.~22, no.~10, pp.
  1345--1359, 2009.

\bibitem{xian2017zero}
Y.~Xian, B.~Schiele, and Z.~Akata, ``Zero-shot learning-the good, the bad and
  the ugly,'' in \emph{Conference on Computer Vision and Pattern Recognition
  (CVPR)}, 2017.

\bibitem{chao2016empirical}
W.-L. Chao, S.~Changpinyo, B.~Gong, and F.~Sha, ``An empirical study and
  analysis of generalized zero-shot learning for object recognition in the
  wild,'' in \emph{European Conference on Computer Vision (ECCV)}, 2016.

\bibitem{DiasDaCruz2020SVIRO}
S.~{Dias Da Cruz}, O.~Wasenm\"uller, H.-P. Beise, T.~Stifter, and D.~Stricker,
  ``Sviro: Synthetic vehicle interior rear seat occupancy dataset and
  benchmark,'' in \emph{IEEE Winter Conference on Applications of Computer
  Vision (WACV)}, 2020.

\bibitem{autopose}
M.~Selim, A.~Firintepe, A.~Pagani, and D.~Stricker, ``Autopose: Large-scale
  automotive driver head pose and gaze dataset with deep head pose baseline,''
  in \emph{International Conference on Computer Vision Theory and Applications
  (VISAPP)}, 2020.

\bibitem{schwarz2017driveahead}
A.~Schwarz, M.~Haurilet, M.~Martinez, and R.~Stiefelhagen, ``Driveahead-a
  large-scale driver head pose dataset,'' in \emph{Conference on Computer
  Vision and Pattern Recognition (CVPR) Workshops}, 2017.

\bibitem{drive_and_act_2019_iccv}
M.~Martin, A.~Roitberg, M.~Haurilet, M.~Horne, S.~Reiß, M.~Voit, and
  R.~Stiefelhagen, ``Drive\&act: A multi-modal dataset for fine-grained driver
  behavior recognition in autonomous vehicles,'' in \emph{International
  Conference on Computer Vision (ICCV)}, 2019.

\bibitem{katrolia2021ticam}
J.~S. Katrolia, B.~Mirbach, A.~El-Sherif, H.~Feld, J.~Rambach, and D.~Stricker,
  ``Ticam: A time-of-flight in-car cabin monitoring dataset,'' 2021.

\bibitem{pub10767}
H.~Feld, B.~Mirbach, J.~Katrolia, M.~Selim, O.~Wasenm{\"u}ller, and
  D.~Stricker, ``Dfki cabin simulator: A test platform for visual in-cabin
  monitoring functions,'' \emph{Commercial Vehicle Technology (CVT) Symposium},
  2021.

\bibitem{nowruzi2019much}
F.~E. Nowruzi, P.~Kapoor, D.~Kolhatkar, F.~A. Hassanat, R.~Laganiere, and
  J.~Rebut, ``How much real data do we actually need: Analyzing object
  detection performance using synthetic and real data,'' \emph{arXiv preprint
  arXiv:1907.07061}, 2019.

\bibitem{tremblay2018training}
J.~Tremblay, A.~Prakash, D.~Acuna, M.~Brophy, V.~Jampani, C.~Anil, T.~To,
  E.~Cameracci, S.~Boochoon, and S.~Birchfield, ``Training deep networks with
  synthetic data: Bridging the reality gap by domain randomization,'' in
  \emph{Conference on Computer Vision and Pattern Recognition (CVPR)}, 2018.

\bibitem{chen2019learning}
Y.~Chen, W.~Li, X.~Chen, and L.~V. Gool, ``Learning semantic segmentation from
  synthetic data: A geometrically guided input-output adaptation approach,'' in
  \emph{Conference on Computer Vision and Pattern Recognition (CVPR)}, 2019.

\bibitem{reiss2020deep}
S.~Rei{\ss}, A.~Roitberg, M.~Haurilet, and R.~Stiefelhagen, ``Deep
  classification-driven domain adaptation for cross-modal driver behavior
  recognition,'' in \emph{IEEE Intelligent Vehicles Symposium (IV)}, 2020.

\bibitem{visapp21}
J.~Katrolia, L.~Kr\"amer, J.~Rambach, B.~Mirbach, and D.~Stricker, ``An
  adversarial training based framework for depth domain adaptation,'' in
  \emph{International Joint Conference on Computer Vision, Imaging and Computer
  Graphics Theory and Applications (VISAPP)}, 2021.

\bibitem{norouzi2013zero}
M.~Norouzi, T.~Mikolov, S.~Bengio, Y.~Singer, J.~Shlens, A.~Frome, G.~S.
  Corrado, and J.~Dean, ``Zero-shot learning by convex combination of semantic
  embeddings,'' \emph{arXiv preprint arXiv:1312.5650}, 2013.

\bibitem{frome2013devise}
A.~Frome, G.~S. Corrado, J.~Shlens, S.~Bengio, J.~Dean, M.~Ranzato, and
  T.~Mikolov, ``Devise: A deep visual-semantic embedding model,'' in
  \emph{Advances in Neural Information Processing Systems (NIPS)}, 2013.

\bibitem{liu2017unsupervised}
M.-Y. Liu, T.~Breuel, and J.~Kautz, ``Unsupervised image-to-image translation
  networks,'' in \emph{Proceedings of the International Conference on Neural
  Information Processing Systems (NIPS)}, 2017.

\bibitem{karras2019style}
T.~Karras, S.~Laine, and T.~Aila, ``A style-based generator architecture for
  generative adversarial networks,'' in \emph{Conference on Computer Vision and
  Pattern Recognition (CVPR)}, 2019.

\bibitem{tsai2018learning}
Y.-H. Tsai, W.-C. Hung, S.~Schulter, K.~Sohn, M.-H. Yang, and M.~Chandraker,
  ``Learning to adapt structured output space for semantic segmentation,'' in
  \emph{Proceedings of the Conference on Computer Vision and Pattern
  Recognition (CVPR)}, 2018.

\bibitem{rangesh2020driver}
A.~Rangesh, B.~Zhang, and M.~M. Trivedi, ``Driver gaze estimation in the real
  world: Overcoming the eyeglass challenge,'' in \emph{IEEE Intelligent
  Vehicles Symposium (IV)}, 2020.

\bibitem{li2017deeper}
D.~Li, Y.~Yang, Y.-Z. Song, and T.~M. Hospedales, ``Deeper, broader and artier
  domain generalization,'' in \emph{International Conference on Computer Vision
  (ICCV)}, 2017.

\bibitem{zhou2020learning}
K.~Zhou, Y.~Yang, T.~Hospedales, and T.~Xiang, ``Learning to generate novel
  domains for domain generalization,'' in \emph{European Conference on Computer
  Vision (ECCV)}, 2020.

\bibitem{VanSteenkiste2019}
S.~van Steenkiste, F.~Locatello, J.~Schmidhuber, and O.~Bachem, ``Are
  disentangled representations helpful for abstract visual reasoning?'' in
  \emph{Advances in Neural Information Processing Systems (NIPS)}, 2019.

\bibitem{burgess2019monet}
C.~P. Burgess, L.~Matthey, N.~Watters, R.~Kabra, I.~Higgins, M.~Botvinick, and
  A.~Lerchner, ``Monet: Unsupervised scene decomposition and representation,''
  \emph{arXiv preprint arXiv:1901.11390}, 2019.

\bibitem{engelcke2019genesis}
M.~Engelcke, A.~R. Kosiorek, O.~P. Jones, and I.~Posner, ``Genesis: Generative
  scene inference and sampling with object-centric latent representations,''
  \emph{arXiv preprint arXiv:1907.13052}, 2019.

\bibitem{johnson2017clevr}
J.~Johnson, B.~Hariharan, L.~van~der Maaten, L.~Fei-Fei, C.~L. Zitnick, and
  R.~Girshick, ``Clevr: A diagnostic dataset for compositional language and
  elementary visual reasoning,'' in \emph{Conference on Computer Vision and
  Pattern Recognition (CVPR)}, 2017.

\bibitem{Gondal2019}
M.~W. Gondal, M.~Wuthrich, D.~Miladinovic, F.~Locatello, M.~Breidt,
  V.~Volchkov, J.~Akpo, O.~Bachem, B.~Sch{\"o}lkopf, and S.~Bauer, ``On the
  transfer of inductive bias from simulation to the real world: a new
  disentanglement dataset,'' in \emph{Advances in Neural Information Processing
  Systems (NIPS)}, 2019.

\bibitem{chen2020simple}
T.~Chen, S.~Kornblith, M.~Norouzi, and G.~Hinton, ``A simple framework for
  contrastive learning of visual representations,'' \emph{arXiv preprint
  arXiv:2002.05709}, 2020.

\bibitem{noroozi2016unsupervised}
M.~Noroozi and P.~Favaro, ``Unsupervised learning of visual representations by
  solving jigsaw puzzles,'' in \emph{European Conference on Computer Vision
  (ECCV)}, 2016.

\bibitem{xie2012image}
J.~Xie, L.~Xu, and E.~Chen, ``Image denoising and inpainting with deep neural
  networks,'' in \emph{Advances in neural information processing systems
  (NIPS)}, 2012.

\bibitem{vincent2010stacked}
P.~Vincent, H.~Larochelle, I.~Lajoie, Y.~Bengio, and P.-A. Manzagol, ``Stacked
  denoising autoencoders: Learning useful representations in a deep network
  with a local denoising criterion,'' \emph{Journal of machine learning
  research}, vol.~11, no. Dec, pp. 3371--3408, 2010.

\bibitem{badrinarayanan2017segnet}
V.~Badrinarayanan, A.~Kendall, and R.~Cipolla, ``Segnet: A deep convolutional
  encoder-decoder architecture for image segmentation,'' \emph{IEEE
  transactions on pattern analysis and machine intelligence}, vol.~39, no.~12,
  pp. 2481--2495, 2017.

\bibitem{bergmann2018improving}
P.~Bergmann, S.~L{\"o}we, M.~Fauser, D.~Sattlegger, and C.~Steger, ``Improving
  unsupervised defect segmentation by applying structural similarity to
  autoencoders,'' \emph{arXiv preprint arXiv:1807.02011}, 2018.

\bibitem{wang2003multiscale}
Z.~Wang, E.~P. Simoncelli, and A.~C. Bovik, ``Multiscale structural similarity
  for image quality assessment,'' in \emph{IEEE Asilomar Conference on Signals,
  Systems and Computers, 2003}, vol.~2, 2003.

\bibitem{hou2017deep}
X.~Hou, L.~Shen, K.~Sun, and G.~Qiu, ``Deep feature consistent variational
  autoencoder,'' in \emph{IEEE Winter Conference on Applications of Computer
  Vision (WACV)}, 2017.

\bibitem{Da_Cruz_2021_WACV}
S.~D. Da~Cruz, B.~Taetz, T.~Stifter, and D.~Stricker, ``Illumination
  normalization by partially impossible encoder-decoder cost function,'' in
  \emph{Proceedings of the IEEE/CVF Winter Conference on Applications of
  Computer Vision (WACV)}, 2021.

\bibitem{imgaug}
A.~B. Jung, K.~Wada, J.~Crall, S.~Tanaka, J.~Graving, C.~Reinders, S.~Yadav,
  J.~Banerjee, G.~Vecsei, A.~Kraft, Z.~Rui, J.~Borovec, C.~Vallentin,
  S.~Zhydenko, K.~Pfeiffer, B.~Cook, I.~Fernández, F.-M. De~Rainville, C.-H.
  Weng, A.~Ayala-Acevedo, R.~Meudec, M.~Laporte, \emph{et~al.}, ``{imgaug},''
  \url{https://github.com/aleju/imgaug}, 2020, online; accessed 01-Feb-2020.

\bibitem{Gongfan2019}
F.~Gongfan, ``Pytorch ms-ssim,'' \url{https://github.com/VainF/pytorch-msssim},
  2019.

\end{thebibliography}

\end{document}